\documentclass[journal]{IEEEtran}
\usepackage{graphicx}
\usepackage{color}
\usepackage[colorlinks=true]{hyperref}
\usepackage{amsmath}
\usepackage{amssymb}
\usepackage{multirow}
\usepackage{algorithm}
\usepackage{algorithmic}
\usepackage{times}
\usepackage{epsfig}
\usepackage{graphicx}
\usepackage{colortbl}
\usepackage{array}

\graphicspath{{Imgs/},{Author/}}

\ifCLASSINFOpdf
\else
\fi
\begin{document}

\title{Discriminative Feature Learning with Foreground Attention for Person Re-identification}

\author{Sanping~Zhou,
        Jinjun~Wang,
        Deyu~Meng,
        Yudong~Liang,
        Yihong~Gong,
        Nanning~Zheng
\thanks{Sanping~Zhou, Jinjun~Wang, Yihong~Gong and Nanning~Zheng are with Institute of Artificial Intelligence and Robotics, Xi'an Jiaotong University, Xi'an, Shaanxi, China. Jinjun~Wang is also with Deep North Inc.}
\thanks{Deyu~Meng is with School of Mathematics and Statistics, Xi'an Jiaotong University, Xi'an, Shaanxi, China.}
\thanks{Yudong~Liang is with School of Computer and Information Technology, Shanxi University, Taiyuan, Shanxi, China.}
\thanks{Corresponding author: Jinjun~Wang. Email: jinjun@mail.xjtu.edu.cn.}}

\markboth{IEEE Transactions on Image Processing}%
{Shell \MakeLowercase{\textit{et al.}}: Bare Demo of IEEEtran.cls for Journals}

\maketitle

\begin{abstract}
The performance of person re-identification~(Re-ID) has been seriously effected by the large cross-view appearance variations caused by mutual occlusions and background clutters. Hence learning a feature representation that can adaptively emphasize the foreground persons becomes very critical to solve the person Re-ID problem. In this paper, we propose a simple yet effective foreground attentive neural network~(FANN) to learn a discriminative feature representation for person Re-ID, which can adaptively enhance the positive side of foreground and weaken the negative side of background. Specifically, a novel foreground attentive subnetwork is designed to drive the network's attention, in which a decoder network is used to reconstruct the binary mask by using a novel local regression loss function, and an encoder network is regularized by the decoder network to focus its attention on the foreground persons. The resulting feature maps of encoder network are further fed into the body part subnetwork and feature fusion subnetwork to learn discriminative features. Besides, a novel symmetric triplet loss function is introduced to supervise feature learning, in which the intra-class distance is minimized and the inter-class distance is maximized in each triplet unit, simultaneously. Training our FANN in a multi-task learning framework, a discriminative feature representation can be learned to find out the matched reference to each probe among various candidates in the gallery. Extensive experimental results on several public benchmark datasets are evaluated, which have shown clear improvements of our method over the state-of-the-art approaches.
\end{abstract}


\IEEEpeerreviewmaketitle

\section{Introduction}
\label{sec_intr}

\IEEEPARstart{P}{erson} re-identification~(Re-ID) is an important task for many surveillance applications such as person association~\cite{Morris_Trivedi:2008}, multi-target tracking~\cite{Zhang_Wang_Wang:2015} and behavior analysis~\cite{Hu_Tan_Wang:2004}. Given a pedestrian image from one camera view, it tries to find out the stated person amongst a set of gallery candidates captured from the disjoint camera views. The person Re-ID problem has attracted extensive research attentions in recent years, and yet it still remains a challenging one due to the large cross-view appearance variations caused by mutual occlusions and background clutters. Therefore, the key to improve the person Re-ID performance is to learn a discriminative feature representation which is robust to the large cross-view appearance variations.

\begin{figure}[!htb]
	\centering
	\begin{tabular}{c}
		\includegraphics[height = 5.2cm, width = 6.7cm]{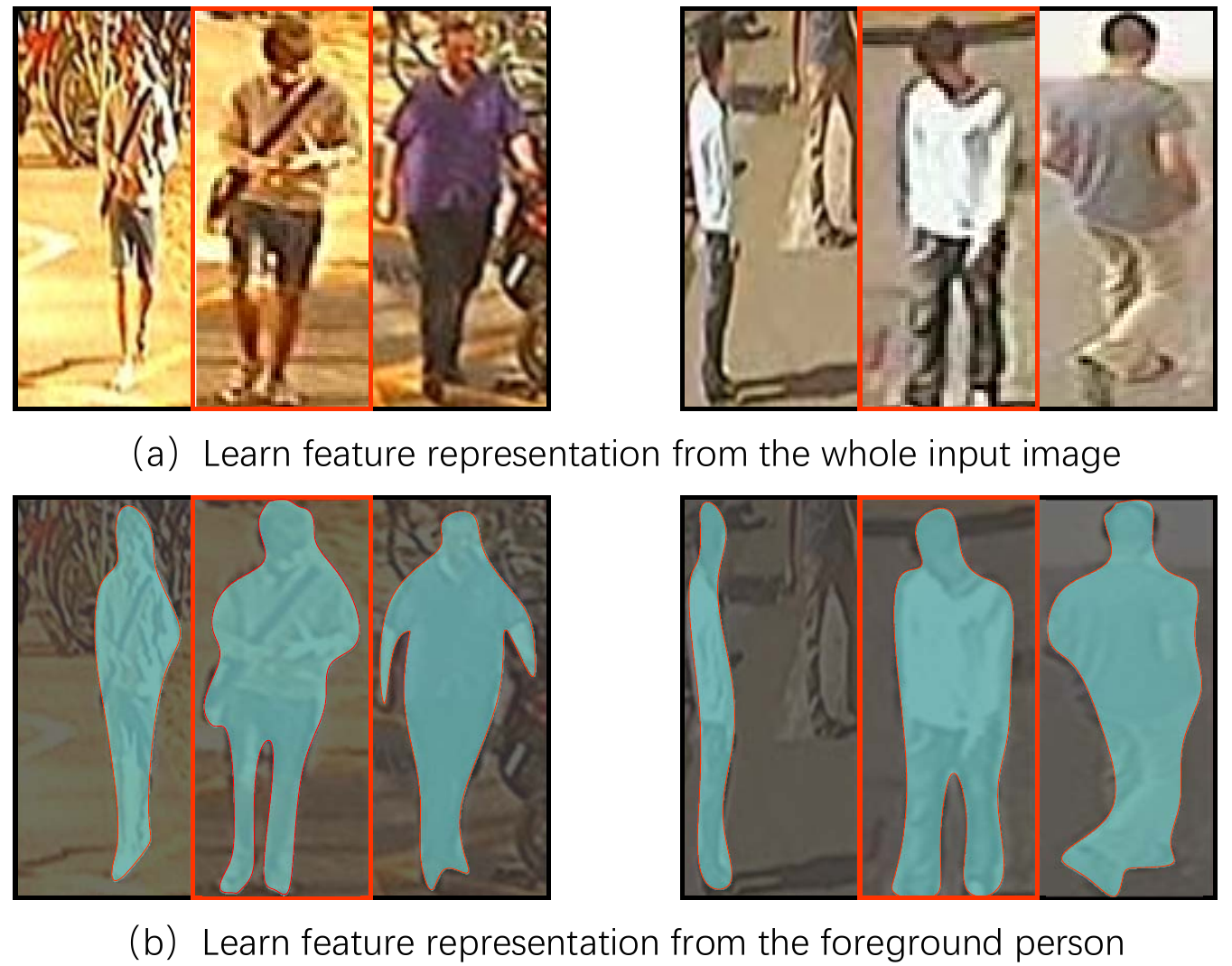}
	\end{tabular}
    \vspace{-0.1cm}
	\caption{Motivation of our method: we aim to learn a discriminative feature representation which can address the foreground of each input image. In particular, row (a) shows that the general way learns feature from the whole image, and row (b) shows that our method learns feature only from the foreground person.}
	\label{fig_1}
	\vspace{-0.3cm}
\end{figure}

To address this problem, extensive works have been reported in the past few years, which could be roughly divided into the following two categories: 1) developing discriminative descriptors to handle the variations of person's appearance, and 2) designing distinctive distance metrics to measure the similarity between images. In the first line of works, different informative feature descriptors have been attempted by utilizing different clues, including the LBP~\cite{Xiong_Gou_Camps:2014}, ELF~\cite{Gray_Tao:2008} and LOMO~\cite{Zhao_Ouyang_Wang:2014}. In the second line of works, labeled images are used to learn effective distance metrics, including the LADF~\cite{Li_Chang_Liang:2013}, LMNN~\cite{Weinberger_Blitzer_Saul:2005} and ITML~\cite{Davis_Kulis_Jain:2007}. An evident drawback of these methods is that they consider feature extraction and metric learning as two independent steps, and therefore they cannot complement their capabilities in a joint framework.

Benefit from the strong representation capacity of deep neural network, the deep feature learning based methods~\cite{Ahmed_Jones_Marks:2015, Ding_Lin_Wang:2015} have significantly improved the person Re-ID results on the public benchmark datasets. These methods are usually consisted of two components, {\em i.e.}, a neural network and an objective function. Specifically, the neural network is built to extract features from input images, and the objective function is designed to guide the training process. Representative deep neural networks include the AlexNet~\cite{Krizhevsky_Sutskever_Hinton:2012}, VGGNet~\cite{Simonyan_Zisserman:2014} and ResNet~\cite{He_Zhang_Ren:2016}, and representative objective functions include the softmax loss function~\cite{Krizhevsky_Sutskever_Hinton:2012}, triplet loss function~\cite{Wang_Song_Leung:2014} and contrastive loss function~\cite{Hadsell_Chopra_LeCun:2006}. These works usually take the entire rectangular images as inputs, therefore
the extracted features may easily get degenerated by the background noises. In order to solve this problem, several works~\cite{Kalayeh_Basaran_Gökmen:2018,Song_Huang_Ouyang:2018,Tian_Yi_Li:2018} have been presented to address the foreground persons in feature learning, which are implemented in two steps: 1) Learn two kinds of features from both the complete and masked images using a multi-path network; and 2) Concatenate the multi-path features and fuse them at the output layers. Because they use a multi-path network to extract features, heavy computations are needed at both the training and testing stages.

In this paper, we incorporate a foreground attentive neural network~(FANN) and a symmetric triplet loss function into an end-to-end feature learning framework,\footnote{Note that we submitted our paper before the available of ~\cite{Kalayeh_Basaran_Gökmen:2018,Song_Huang_Ouyang:2018,Tian_Yi_Li:2018}, therefore it can be viewed a co-occurring piece of work. Besides, the symmetric triplet loss function was originally proposed in~\cite{Zhou_Wang_Wang:2017}, and this work is an extension of our conference paper.} so as to learn a discriminative feature representation from the foreground images for person Re-ID. As illustrated in Fig.~\ref{fig_1}, a detected person in a rectangular image region may easily include background clutters and mutual occlusions from the other objects. If such noises can be attenuated, the extracted features will mainly come from the foreground persons, which are more discriminative and robust to the large cross-view appearance variations. This observation has motivated us to propose a novel multi-task learning framework that can jointly alleviate the side effects of backgrounds and learn the discriminative features from foregrounds. Specifically, a novel FANN is built to focus its attention on the foreground persons, in which each image is first passed through an encoder and decoder network, then the outputs of encoder network are further taken for the discriminative feature learning. The encoder network extracts features from the whole image, and the decoder network reconstructs a binary mask of each foreground person. As a result, the encoder network will gradually focus its attention on the foreground persons with the regularization of decoder network by using a novel local regression loss function. Besides, a novel symmetric triplet loss function is introduced to learn the discriminative features, in which the intra-class distance is minimized and the inter-class distance is maximized in each triplet unit, simultaneously. Training the FANN in an end-to-end manner, the foreground attentive features can be finally learned to distinguish different individuals across the disjoint camera views. Extensive experimental results on the 3DPeS~\cite{Baltieri_Vezzani_Cucchiara:2011}, VIPeR~\cite{Gray_Tao:2008}, CUHK01~\cite{Li_Wang:2013}, CUHK03~\cite{Li_Zhao_Xiao:2014}, Market1501~\cite{Zheng_Shen_Tian:2015} and DukeMTMC-reID~\cite{Ristani_Solera_Zou:2016} datasets have shown the significant improvements by our method, as compared with the state-of-the-art approaches.

The main contributions of this work can be highlighted as follows:
\begin{itemize}
  \item We design a simple yet effective FANN to learn robust features for person Re-ID, in which the side effects of background can be naturally attenuated and the useful clues in foreground can be greatly emphasized.
  \item We build an effective local regression loss function to supervise the foreground mask reconstruction, in which the local information in a small neighborhood is used to smooth the isolated regions in ground truth mask.
  \item We introduce a novel symmetric triplet loss function to supervise the feature learning, in which the intra-class distance is minimized and the inter-class is maximized in each triplet unit, simultaneously.
\end{itemize}

The rest of our paper is organized as follows: In Section~\ref{sec_rel}, we briefly review the related works. Section~\ref{sec_alog} introduces our neural network and objective function, followed by a discussion of the learning algorithm in Section~\ref{sec_opt}. Experimental results and ablation studies are presented in Section~\ref{sec_exp}. And conclusion comes in Section~\ref{sec_con}.
\section{Related Work}
\label{sec_rel}
We review three lines of related works, including the metric learning based method, the deep learning based method and the attention learning based method, which are briefly introduced in the following paragraphs.

{\bf Metric learning based method.} This category of methods aim to find a mapping function from the feature space to distance space, in which distances between images of the same person are closer than those between different identities. For example, Zheng et al.~\cite{Zheng_Gong_Xiang:2013} proposed a relative distance learning method from the probabilistic prospective. In~\cite{Mignon_Jurie:2012}, Mignon et al. learned a distance metric with the sparse pairwise similarity constraints. Pedagadi et al.~\cite{Pedagadi_Orwell_Velastin:2013} utilized the Local Fisher Discriminant Analysis~(LFDA) to map the high dimensional features into a more discriminative low dimensional space. In ~\cite{Xiong_Gou_Camps:2014}, Xiong et al. further extended the LFDA and several other metrics by using the kernel tricks and different regularizers. Nguyen et al.~\cite{Nguyen_Bai:2011} measured the similarity between image pairs through the cosine similarity, which was closely related to the inner product similarity. In~\cite{Loy_Liu_Gong:2013}, Loy et al. casted the person Re-ID problem as an image retrieval task by considering the listwise similarity. Chen et al.~\cite{Chen_Yuan_Hua:2015} proposed a kernel based metric learning method to explore the nonlinear relationship of samples in feature space. In~\cite{Hirzer_Roth:2012}, Hirzer et al. learned a discriminative metric by using the relaxed pairwise constraints. These methods try to learn a specific distance metric based on features extracted from the fixed feature descriptors, which could not fully discover the potential of metric learning.

\begin{figure*}[!htb]
	\centering
	\begin{tabular}{c}
		\includegraphics[height = 6.5cm, width = 17.5cm]{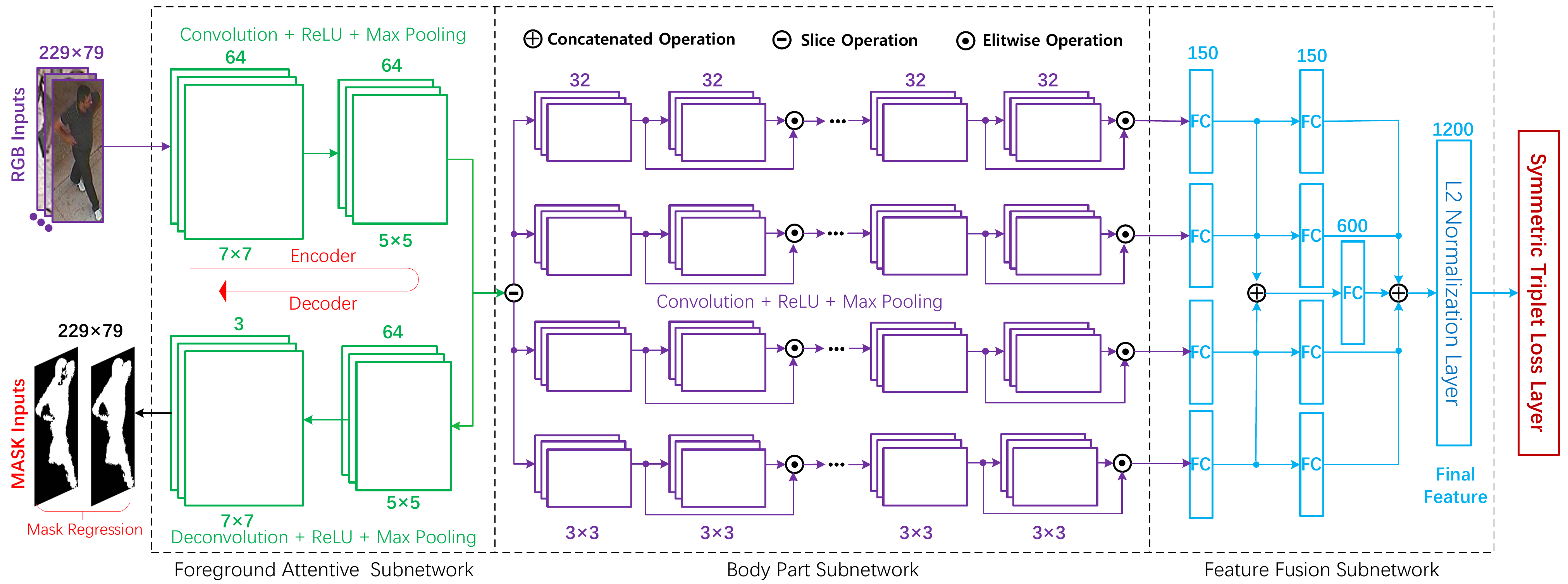}
	\end{tabular}
    \vspace{-0.1cm}
	\caption{Illustration of our deep neural network, which is consisted of the foreground attentive subnetwork, the body part subnetwork and the feature fusion subnetwork. Specifically, the foreground attentive subnetwork aims to focus the attention on foreground by passing each input image through an encoder and decoder network. Then, the encoded feature maps are averagely sliced and discriminately learned in the following body part subnetwork. Afterwards, the resulting feature maps are fused in the feature fusion subnetwork. Finally, the final feature vectors are normalized to a unit sphere space and learned by following the symmetric triplet loss layer.}
	\label{fig_2}
	\vspace{-0.3cm}
\end{figure*}

{\bf Deep learning based method.} This category of methods usually incorporate feature extraction and metric learning into a joint framework, in which a neural network is used to extract features and a distance metric is used to compute losses and back-propagate gradients. For example, Ahmed et al.~\cite{Ahmed_Jones_Marks:2015} proposed a novel deep neural network which took the pairwise images as inputs, and output a similarity value indicating whether two input images were the same person or not. In~\cite{Xiao_Li_Ouyang:2016}, Xiao et al. applied a domain guided dropout algorithm to learn the general features for person Re-ID. Ding et al.~\cite{Ding_Lin_Wang:2015} introduced a triplet neural network to learn the relative similarity in solving the person Re-ID problem. In~\cite{Wang_Zuo_Lin:2016}, Wang et al. proposed an unified triplet and siamese deep architecture, which could jointly extract the single-image and cross-image feature representation. Zhou et al.~\cite{Zhou_Huang_Wang:2017} applied a recurrent neural network to jointly learn the spatial and temporal features from video sequence. In~\cite{Shen_Li_Xiao:2018}, Shen et al. designed a novel group-shuffling random walk network for fully utilizing the affinity information between gallery images in both the training and testing stages. Xiao et al.~\cite{Xiao_Li_Wang:2017} proposed an unified framework which can jointly handle the pedestrian detection and person Re-ID in a single network. One major limitation of these methods is that they take the whole image as input, which isn't able to extensively address the foreground persons. Therefore, the learned features will be easily effected by the background noises.

{\bf Attention learning based method.} This category of methods aim at learning a discriminative feature representation from input images by using different attention mechanisms~\cite{Jaderberg_Simonyan_Zisserman:2015}, which can be roughly divided into two categories, {\em i.e.}, the supervised and unsupervised approaches. In the former ones, the ground truth is needed to supervise the attention learning. For example, Kalayeh et al.~\cite{Kalayeh_Basaran_Gkmen:2018} took the human parsing results to guide the feature learning for person Re-ID. In~\cite{Song_Huang_Ouyang:2018}, Song et al. designed a mask-guided network to drive the network's attention on the foregrounds of input images, which was effective to learn features from the discriminative body regions. Meanwhile, Tian et al.~\cite{Tian_Yi_Li:2018} also studied how to alleviate the side effect of background in feature learning. In the latter ones, the attention learning process is usually driven by a specific task or regularizer, which is less effective because no labeled information is available. For example, Wang et al.~\cite{Wang_Jiang_Qian:2017} proposed a residual attention network which embedded an attention mechanism in the network for image classification. In~\cite{Zhao_Li_Zhuang:2017}, Zhao et al. proposed a part-aligned representation learning method to aggregate the similarities between the corresponding regions of person images. Li et al.~\cite{Li_Zhu_Gong:2018} designed a harmonious attention network to jointly learn the soft pixel attention and hard regional attention for person Re-ID. In~\cite{Gheissari_Sebastian_Hartley:2006}, Gheissari et al. introduced a novel spatial-temporal segmentation algorithm to generate the salient regions for person Re-ID. The supervised methods are usually more expensive but effective than the unsupervised ones. In order to pursue higher accuracy, we presented a novel supervised attention learning method to learn discriminative features from the foreground persons, in which the regression task is designed to regularize the feature learning by gradually reconstructing the foreground masks in the training process. Therefore, the complexity of feature extraction network will not be increased in the testing phase, as compared with the existing attention learning based methods in person Re-ID.

\section{Multi-task framework for foreground attentive feature learning}
\label{sec_alog}
\subsection{Foreground Attentive Neural Network}
The goal of our FANN is to learn a discriminative feature representation from the foregrounds of input images. The proposed network is shown in Fig.~\ref{fig_2}, which is consisted of the foreground attentive subnetwork, the body part subnetwork and the feature fusion subnetwork. The details are explained in the following paragraphs.

{\bf Foreground attentive subnetwork.} The foreground attentive subnetwork aims to focus its attention on the foregrounds of input images, so as to alleviate the side effects of backgrounds. Our adopted paradigm is to pass each input image through an encoder and decoder network, in which the encoder network extracts features from the RGB images and the decoder network reconstructs the binary mask of foreground person. The encoder network will naturally focus its attention on the foreground, since the decoder network can gradually reconstruct the binary foreground mask in the learning process. Specifically, the input images are first resized to $229\times79$ and passed through two $64$ learned filters in size of $7\times7$ and $5\times5$ with strides $3$ and $2$, respectively. Then, the resulting feature maps are passed through a rectified linear unit~(ReLU) and followed by a max pooling kernel in size of $3\times3$ with stride $1$. These layers constitute the encoder network, and the outputs of encoder network are further fed into the decoder network and the body part subnetwork, simultaneously. The decoder network is consisted of two deconvolutional layers, which are with $64$ and $3$ learned filters in size of $5\times5$ and $7\times7$ with strides $2$ and $3$, respectively. In addition, a rectified linear unit~(ReLU) is put between the two layers. The output of decoder network is used to reconstruct the binary mask of foreground person, so as to adaptively drive the attention of encoder network on the foregrounds of input images.

\begin{figure}[t]
	\centering
	\begin{tabular}{c}
		\hspace{-0.2cm}
		\includegraphics[height = 6.0cm, width = 8.5cm]{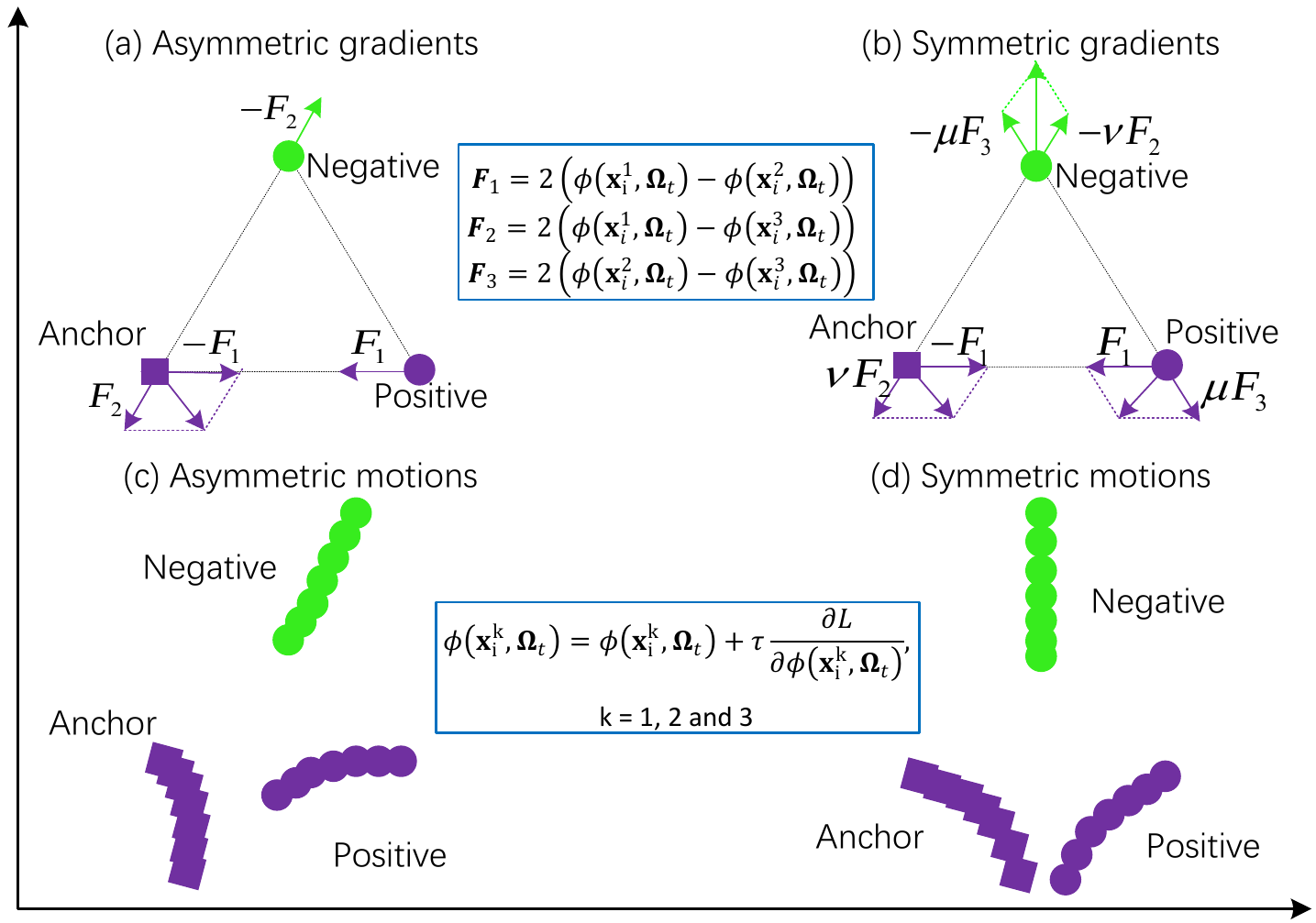}
	\end{tabular}
    \vspace{-0.1cm}
	\caption{Illustration of gradient back-propagations and motion trajectories driven by two different triplet loss functions. Specifically, (a) shows the gradients of asymmetric triplet loss function; (b) shows the gradients of symmetric triplet loss function, (c) shows the motion trajectory driven by asymmetric triplet loss function, and (d) shows the motion trajectory driven by symmetric triplet loss function. In the optimization process, we adaptively update $u$ and $v$, so as to jointly minimize the intra-class distance and maximize the inter-class distance in each triplet unit.}
	\label{fig_3}
	\vspace{-0.3cm}
\end{figure}

{\bf Body part subnetwork.} The body part subnetwork aims at learning a discriminative feature representation from different body parts, which is inspired by the idea that different body parts have different weights in representing one person~\cite{Ahmed_Jones_Marks:2015}. The resulting feature maps of encoder network are first averagely sliced into four equal parts across the height channel, and then the sliced feature maps are fed into the body part subnetwork for feature learning. The body part subnetwork is consisted of four sets of residual blocks~\cite{He_Zhang_Ren:2016}, in which these convolutional layers do not share parameters, so as to discriminatively learn feature representations from different body parts. In each residual block, we pass each set of sliced feature maps through two small convolutional layers, in which both of them have $32$ learned filters in size of $3\times3$ with stride $1$. The outputs of first small convolutional layer are summarized with the outputs of second small convolutional layer by using the eltwise operation. Then, a rectified linear unit~(ReLU) is followed after them. Finally, the resulting feature maps are passed through a max pooling kernel in size of $3\times3$ with stride $1$. In order to enhance the feature representation capacity, we add several residual blocks after the first one, and all of them are in the same shape. Notice that the actual number should be determined by the scale of training dataset. 

{\bf Feature fusion subnetwork.} The feature fusion subnetwork aims to fuse the learned features and normalize them to a unit sphere space. It is consisted of four teams of fully connected layers and a $\mathrm{L}2$ normalization layer. Specifically, the local feature maps of each body part are first discriminately learned by following two small fully connected layers in each team. The dimensions of these small fully connected layers are $150$. Then, a rectified linear unit~(ReLU) is added between them. Afterwards, the discriminatively learned features of the first four small fully connected layers are concatenated to be fused by following a large fully connected layer, whose dimension is $600$. Finally, the resulting feature vectors are further concatenated with the outputs of second four fully connected layers, so as to generate the final $1200$ dimensional feature vectors for representation. In addition, a $\mathrm{L}2$ normalization layer is used to regularize the magnitude of each feature vector to be unit. Therefore, the similarity comparison measured in the Euclidian distance is equivalent to that by using the cosine distance.

\subsection{Multi-Task Objective Function}
Let $\mathbf{Y} = \left\{ \mathbf{X}_i, \mathbf{M}_i\right\}_{i=1}^N$ be the input training data, in which $\mathbf{X}_i$ denotes the RGB image, $\mathbf{M}_i$ represents the mask of foreground, and $N$ is the number of training samples. Specifically, $\mathbf{X}_i = \{\mathbf{x}_i^1, \mathbf{x}_i^2, \mathbf{x}_i^3\}$ indicates the $i^{th}$ triplet unit, in which $\mathbf{x}_i^1$ and $\mathbf{x}_i^2$ are two images with the same identity, $\mathbf{x}_i^1$ and $\mathbf{x}_i^3$ are two mismatched images with different identities. Besides, $\mathbf{M}_i = \{\mathbf{m}_i^1, \mathbf{m}_i^2, \mathbf{m}_i^3\}$ represents the corresponding foreground mask of $\mathbf{X}_i$. The goal of our FANN is to learn filter weights and biases that can jointly minimize the ranking error and the reconstruction error at the output layers, respectively. A recursive function for an $M$-layer deep model can be defined as follows:
\begin{equation}
\label{eq_1}
\begin{aligned}
\mathbf{Y}_i^{l} &= \phi(\mathbf{W}^{l}*\mathbf{Y}_i^{l-1} + \mathbf{b}^{l}) \\
i = 1, \cdots , &N; l = 1, \cdots, M; \mathbf{Y}_i^{0} = \mathbf{Y}_i.
\end{aligned}\,,
\end{equation}
{\noindent
where $\mathbf{W}^{l}$ denotes the filter weights of the $l^{th}$ layer, $\mathbf{b}^{l}$ refers to the corresponding biases, $*$ denotes the convolution operation, ${\phi}(\cdot)$ is an element-wise non-linear activation function such as ReLU, and $\mathbf{Y}_i^{l}$ represents the feature maps generated at layer $l$ for $\mathbf{Y}_i$. For simplicity, we consider the deep parameters as a whole $\mathbf{\Omega}=\{\mathbf{W},\mathbf{b}\}$, in which $\mathbf{W} = \{\mathbf{W}^{1},\cdots, \mathbf{W}^{M}\}$ and $\mathbf{b} = \{\mathbf{b}^{1},\cdots, \mathbf{b}^{M}\}$}.

In order to train our FANN in an end-to-end manner, we apply a multi-task objective function to supervise the learning process, which is defined as follows:
\begin{equation}
\label{eq_2}
\begin{aligned}
&\mathop{\min} \limits_{\mathbf{\Omega}, \mathbf{u}, \mathbf{v}} {\rm E}(\mathbf{\Omega}, \mathbf{u}, \mathbf{v}) =\\
&\sum \limits_{i = 1}^N  \mathrm{L}_1(u_i, v_i, \phi(\mathbf{X}_i, \mathbf{\Omega}_t)) + \zeta \mathrm{L}_2(\phi(\mathbf{X}_i, \mathbf{\Omega}_r), \mathbf{M}_i) + \eta \mathrm{R}(\mathbf{\Omega}),
\end{aligned}
\end{equation}
where $\mathrm{L}_1(\cdot)$ denotes the symmetric triplet loss term, $\mathrm{L}_2(\cdot)$ represents the local regression term, $\mathrm{R}(\cdot)$ indicates the parameter regularization term, and $\zeta, \eta$ are two fixed weight parameters. Specifically, $\mathbf{u} = [u_1,\dots,u_N]$ and $\mathbf{v} = [v_1,\dots,v_N]$ are two adaptive weights which control the symmetric gradient back-propagation. Besides, $\mathbf{\Omega} = [\mathbf{\Omega}_t, \mathbf{\Omega}_r]$, in which $\mathbf{\Omega}_t$ is the parameters of deep ranking network and $\mathbf{\Omega}_r$ is the parameters of deep regression network.

{\bf Symmetric triplet loss term.} The goal of our symmetric triplet loss function is to jointly minimize the intra-class distance and maximize the inter-class distance in each triplet unit, so as to learn a discriminative feature representation to correctly match images of each individual captured from the disjoint camera views. Its superiority against the asymmetric triplet loss function~\cite{Wang_Song_Leung:2014} is that the deduced gradients to the positive samples are symmetric, as shown in Fig.~\ref{fig_3}, which is very essential to consistently minimize the intra-class distance in the training process~\footnote{The reason of why our symmetric triplet loss function outperforms the asymmetric one is that it can accelerate the motion of positive samples in the vertical direction, as shown in Fig.~\ref{fig_3}. Therefore, the intra-class distance can be consistently minimized in the training process.}. The hinge loss of our symmetric triplet loss function is formulated as follows:
\begin{equation}
\label{eq_3}
\begin{aligned}
\mathrm{L}_1 = \max \{\mathrm{M} + \mathrm{d}(\mathbf{x}_i^1,\mathbf{x}_i^2) - [\mathrm{u}_i \mathrm{d}(\mathbf{x}_i^1,\mathbf{x}_i^3) + \mathrm{v}_i \mathrm{d}(\mathbf{x}_i^2,\mathbf{x}_i^3)],0\}
\end{aligned}\,,
\end{equation}
where $\mathrm{M}$ is the margin between the positive pair and negative pair, and $\mathrm{d}(\cdot)$ denotes the pairwise distance measured in the unit spherical space, which is defined as follows:
\begin{equation}
\label{eq_4}
\begin{aligned}
\mathrm{d}(\mathbf{x}_i^j,\mathbf{x}_i^k) = \|\phi(\mathbf{x}_i^j, \mathbf{\Omega}_t) - \phi(\mathbf{x}_i^k, \mathbf{\Omega}_t)\|_2^2.
\end{aligned}
\end{equation}

{\noindent
In practice, we need to normalize $\|\phi(\mathbf{x}_i^j, \mathbf{\Omega}_t)\|_2^2 = 1$ and $\|\phi(\mathbf{x}_i^k, \mathbf{\Omega}_t)\|_2^2 = 1$, therefore the distance measured in the Euclidean space is equivalent to that measured in the unit spherical space. The smaller the distance $\mathrm{d}(\mathbf{x}_i^j,\mathbf{x}_i^k)$ is, the more similar the two input images $\mathbf{x}_i^j$ and $\mathbf{x}_i^k$ are, and vice versa. Notice that the improved triplet loss function~\cite{Zhang_Gong_Wang:2016} is also lack in ability to deduce the symmetric gradients to positive pairs, because it can't keep $\mathrm{d}(\mathbf{x}_i^1,\mathbf{x}_i^3) \approx \mathrm{d}(\mathbf{x}_i^2,\mathbf{x}_i^3)$ in the training process. In the optimization section, we will explain the underling reason in detail.}

\begin{figure}[!htb]
	\centering
	\begin{tabular}{c}
		\includegraphics[height = 3.5cm, width = 7.5cm]{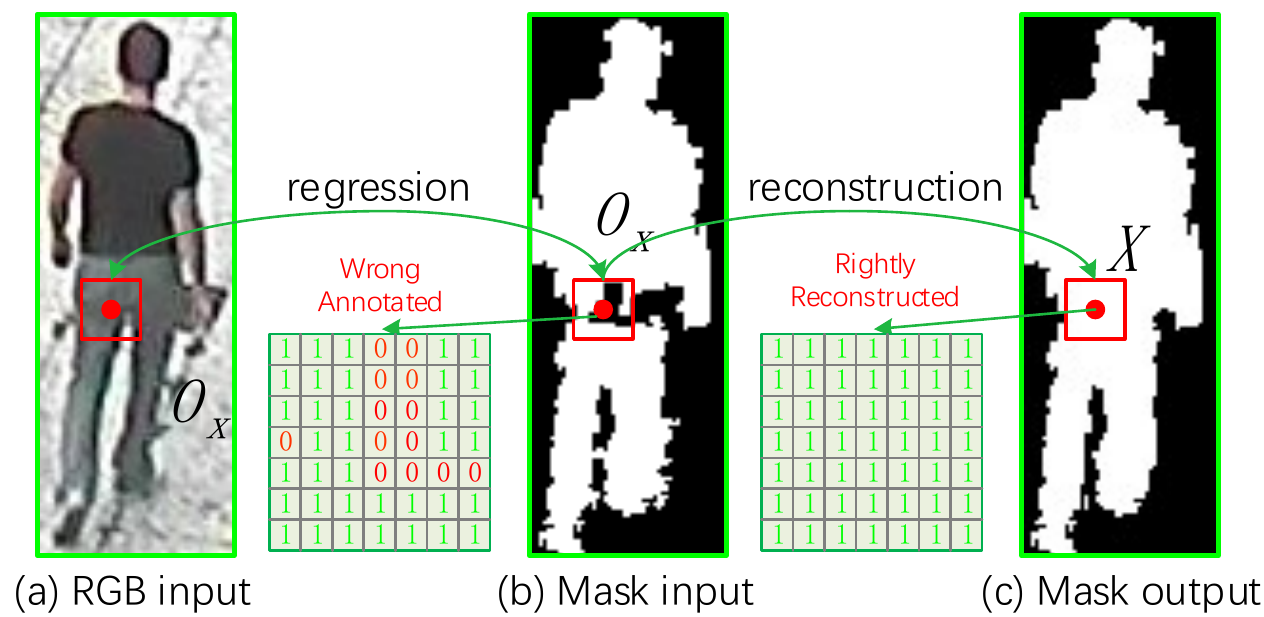}
	\end{tabular}
    \vspace{-0.1cm}
	\caption{Illustration of the binary mask reconstruction in a local neighborhood. In practice, some wrongly annotated foreground pixels can be properly rectified by considering the reconstruction in a local neighborhood.}
	\vspace{-0.3cm}
	\label{fig_4}
\end{figure}

{\bf Local regression loss term.} The goal of our local regression loss function is to minimize the reconstruction error at the output of decoder network. As a result, the encoder network will be regularized by the decoder network in reconstructing the binary masks, and the attention of encoder network can be gradually focused on the foreground persons. We measure the reconstruction error of each pixel in a local neighborhood, which is formulated as follows:
\begin{equation}
\label{eq_5}
\mathrm{L}_2 = \sum \limits_{k = 1}^3  \|K_\sigma * (\phi(\mathbf{x}_i^k,\mathbf{\Omega}_r)-\mathbf{m}_{i}^k)\|_F^2,
\end{equation}
where $K_\sigma$ represents a truncated Gaussian kernel with the standard deviation of $\sigma$, which is formulated as follows:
\begin{equation}
\label{eq_6}
K_\sigma(x-y) = \left\{
\begin{split}
    \frac{1}{\sqrt 2\pi \sigma}\exp (- \frac{|x - y|^2}{2 \sigma ^2} )&, if \hspace{0.05cm}|x \hspace{-0.05cm}-\hspace{-0.05cm} y| \le \rho, \\0{\kern 40pt}&,\hspace{0.05cm}else.
\end{split} \right.,
\end{equation}
where $\rho$ indicates the radius of local neighborhood ${O_x}$ which is centered at the point of $x$. By considering the reconstruction problem in a local neighborhood, the final performance is more robust to the poor mask annotation. As shown in Fig.~\ref{fig_4}, some pixels in the foreground are wrongly labeled as background, and the reconstruction accuracy will be seriously effected if we reconstruct the foreground mask by only measuring the point to point difference. In our method, we measure the point to set difference by jointly considering the neighborhood information, therefore the foreground mask will be properly reconstructed if most of the pixels in a local neighborhood can be rightly annotated.

{\bf Parameter regularization term.} The goal of our parameter regularizer is to smooth the parameters of the entire neural network, which is formulated as follows:
\begin{equation}
\label{eq_7}
\mathrm{R} = \sum\limits_{l = 1}^M \| \mathbf{W}^l\|_F^2 + \| \mathbf{b}^l\|_2^2\,,
\end{equation}
where $\|\cdot\|_F^2$ indicates the Frobenius norm, and $\|\cdot\|_2^2$ denotes the Euclidian norm.
\section{Optimization}
\label{sec_opt}
We apply the momentum method to optimize the direction control weights and the stochastic gradient descent algorithm to optimize the deep parameters, which are introduced in the following paragraphs.

The weight parameters $u_i$ and $v_i$ can be adaptively updated in the training process by using the momentum method, so as to jointly minimize the intra-class distance and maximize the inter-class distance in each triplet unit. In order to simplify this problem, we define $u_i = \alpha_i + \beta_i$ and $v_i = \alpha_i - \beta_i$, and therefore the two parameters can be optimized by only updating $\beta_i$ in each iteration. The partial derivative of our symmetric triplet loss function with respect to $\beta_i$ can be formulated as follows:
\begin{equation}
\label{eq_8}
     t = \left\{
    \begin{array}{l}
        \frac{\partial \mathrm{T}(\mathbf{x}_i^1,\mathbf{x}_i^2, \mathbf{x}_i^3)}{\partial \beta_i}, \hspace{0.1cm} if \hspace{0.1cm} \mathrm{T} > 0\,,\\
        \hspace{0.9cm}0\hspace{0.7cm}, \hspace{0.1cm}else.
    \end{array} \right.,
\end{equation}
where $\mathrm{T}= \mathrm{M}+ \mathrm{d}(\mathbf{x}_i^1,\mathbf{x}_i^2) - [u_i \mathrm{d}(\mathbf{x}_i^1,\mathbf{x}_i^3) + v_i \mathrm{d}(\mathbf{x}_i^2,\mathbf{x}_i^3)]$, and $\frac{\partial \mathrm{T}}{\partial \beta_i}$ is formulated as follows:
\begin{equation}
\label{eq_9}
 \frac{\partial \mathrm{T}}{\partial \beta_i} = \|\phi(\mathbf{x}_i^2, \mathbf{\Omega}_t)-\phi(\mathbf{x}_i^3, \mathbf{\Omega}_t)\|_2^2 - \|\phi(\mathbf{x}_i^1, \mathbf{\Omega}_t)-\phi(\mathbf{x}_i^3, \mathbf{\Omega}_t)\|_2^2.
\end{equation}
Then, $\beta_i$ can be optimized as follows:
\begin{equation}
\label{eq_10}
\begin{aligned}
 \beta_i \leftarrow \beta_i - \gamma \cdot t,
\end{aligned}
\end{equation}
where {\small $\gamma$} is the weight updating rate. It can be clearly seen that when $\|\phi(\mathbf{x}_i^1, \mathbf{\Omega}_t)-\phi(\mathbf{x}_i^3, \mathbf{\Omega}_t)\|_2^2 > \|\phi(\mathbf{x}_i^2, \mathbf{\Omega}_t)-\phi(\mathbf{x}_i^3, \mathbf{\Omega}_t)\|_2^2$, namely $t < 0$, then $u_i$ will be decreased while $v_i$ will be increased; and vice verse. As a result, the strength of back-propagation to samples in each triplet unit will be adaptively tuned, in which the anchor and the positive will be clustered, and the negative one will be far away from the hyper-line expanded by the anchor and the positive. Without this property, the improved triplet loss function~\cite{Zhang_Gong_Wang:2016} can not consistently minimize the intra-class distance in the training process.

In order to apply the stochastic gradient descent algorithm to optimize the deep parameters, we compute the partial derivative of our objective function as follows:
\begin{equation}
\label{eq_11}
    \frac{\partial \mathrm{E}}{\partial \mathbf{\Omega}} \hspace{-0.1cm} = \hspace{-0.1cm}\sum\limits_{i = 1}^N \hspace{-0.1cm}\ell_1(u_i, v_i, \phi(\mathbf{X}_i, \mathbf{\Omega}_t)\hspace{-0.05cm}) + \zeta \ell_2(\phi(\mathbf{X}_i, \mathbf{\Omega}_r),\mathbf{M}_i) + \eta \hspace{-0.05cm}\sum\limits_{l = 1}^M \mathbf{\Omega}^l,
\end{equation}
where the first term represents the gradient of symmetric triplet loss function, the second term denotes the gradient of local regression loss function, and the third term indicates the gradient of parameter regularizer.

By the definition of $\mathrm{T}(\mathbf{x}_i^1,\mathbf{x}_i^2,\mathbf{x}_i^3)$ in Eq.~\eqref{eq_8}, the gradient of our symmetric triplet loss term can be computed as follows:
\begin{equation}
\label{eq_12}
     \ell_1 = \left\{ \begin{array}{l}
    \frac{{\partial \mathrm{T}(\mathbf{x}_i^1,\mathbf{x}_i^2,\mathbf{x}_i^3)}}{{\partial \mathbf{\Omega}_t}}, \hspace{0.05cm} if \hspace{0.05cm} \mathrm{T}>0,\\
    \hspace{0.9cm}0\hspace{0.7cm},\hspace{0.05cm} else.
\end{array} \right.,
\end{equation}
where $\frac{\partial \mathrm{T}}{{\partial \mathbf{\Omega}_t}}$ is formulated as follows:
\begin{equation}
\label{eq_13}
\begin{aligned}
   \frac{{\partial \mathrm{T}}}{{\partial \mathbf{\Omega}_t}} = 2(\phi(\mathbf{x}_i^1,  \mathbf{\Omega}_t) - \phi(\mathbf{x}_i^2, \mathbf{\Omega}_t))'\frac{{\partial \phi(\mathbf{x}_i^1, \mathbf{\Omega}_t) - \partial \phi(\mathbf{x}_i^2, \mathbf{\Omega}_t)}}{{\partial \mathbf{\Omega}_t}}&\\
-2u_i(\phi(\mathbf{x}_i^1, \mathbf{\Omega}_t) - \phi(\mathbf{x}_i^3, \mathbf{\Omega}_t))'\frac{{\partial \phi(\mathbf{x}_i^1, \mathbf{\Omega}_t)- \partial \phi(\mathbf{x}_i^3, \mathbf{\Omega}_t)}}{{\partial \mathbf{\Omega}_t}}&\\
-2v_i(\phi(\mathbf{x}_i^2, \mathbf{\Omega}_t) -\phi(\mathbf{x}_i^3, \mathbf{\Omega}_t))'\frac{{\partial \phi(\mathbf{x}_i^2, \mathbf{\Omega}_t)- \partial \phi(\mathbf{x}_i^3, \mathbf{\Omega}_t)}}{{\partial \mathbf{\Omega}_t}}&
\end{aligned}.
\end{equation}

According to the definition of our local regression loss term in Eq.~\eqref{eq_5}, the gradient can be computed as follows:
\begin{equation}
\label{eq_14}
\begin{aligned}
   \hspace{-0.1cm}\ell_2 = \sum\limits_{k = 1}^3 2{K_\sigma} * ({K_\sigma} * (\phi(\mathbf{x}_i^k, \mathbf{\Omega}_r)-\mathbf{m}_i^k)) \frac{\partial \phi(\mathbf{x}_i^k, \mathbf{\Omega}_r)}{\partial \mathbf{\Omega}_r}
\end{aligned}.
\end{equation}
It is clear that the gradients of samples can be easily calculated given the values of $\phi(\mathbf{x}_i^k, \mathbf{\Omega}_t)$, ${\partial \phi(\mathbf{x}_i^k, \mathbf{\Omega}_t)}/{\partial \mathbf{\Omega}_t}$ and $\phi(\mathbf{x}_i^k, \mathbf{\Omega}_r)$, ${\partial \phi(\mathbf{x}_i^k, \mathbf{\Omega}_r)}/{\partial \mathbf{\Omega}_r}$ in each mini-batch, which can be easily obtained by running the forward and backward propagation in the training process. As the algorithm needs to back-propagate the gradients to learn a foreground attentive feature representation, we call it the foreground attentive gradient descent algorithm. Algorithm~\ref{alg} shows the overall process of our implementation regime.

\begin{algorithm}[tb]
   \caption{Foreground Attentive Gradient Descent.}
   \label{alg}
\begin{algorithmic}
   \STATE {\bfseries Input:} \\
   \hspace{0.5 cm}Training data $\mathbf{Y}$, learning rate $\tau$, maximum iterative number ${\rm H}$, weight parameters $\zeta,\eta$, kernel parameters $\sigma, \rho$, margin parameter $\mathcal{M}$, initial weights of $u_i$ and $v_i$ and updating rate $\gamma$.
   \STATE {\bfseries Output:}\\
   \hspace{0.5 cm}The network parameters $\mathbf{\Omega} = [\mathbf{\Omega}_t, \mathbf{\Omega}_r]$.
   \REPEAT
   \STATE 1. Extract the features of $\phi(\mathbf{x}_i^k, \mathbf{\Omega}_t)$ and $\phi(\mathbf{x}_i^k, \mathbf{\Omega}_r)$ in each triplet unit by the forward propagation.
   \REPEAT
   \STATE \hspace{-0.2cm}a) Compute the gradient of $\frac{\partial \mathrm{T}}{\partial \beta_i}$ according to Eq.~\eqref{eq_9};
   \STATE \hspace{-0.2cm}b) Update weights $u_i$ and $v_i$ according to Eq.~\eqref{eq_10};
   \STATE \hspace{-0.2cm}c) Compute the gradients of $\ell_1$ and $\ell_2$ according to Eq.~\eqref{eq_12} and Eq.~\eqref{eq_14};
   \STATE \hspace{-0.2cm}d) Update the gradients of $\frac{\partial \mathrm{E}}{\partial \mathbf{\Omega}}$ according to Eq.~\eqref{eq_11};
   \UNTIL{Traverse all the triplet inputs $\{{\mathbf{y}_i^1,\mathbf{y}_i^2,\mathbf{y}_i^3}\}$ in each min-batch};
   \STATE 2. Update $\mathbf{\Omega}^{(h+1)} = \mathbf{\Omega}^{(h)} - {\tau_h}\frac{\partial {\rm E}}{\partial \mathbf{\Omega}^{(h)}}$ and $h \leftarrow h + 1$.
   \UNTIL{$h > {\rm H}$}
\end{algorithmic}
\end{algorithm}

\section{Experiments}
\label{sec_exp}

\subsection{Datasets and Settings}
{\bf Benchmark datasets.} We evaluate our method on six datasets, including the 3DPeS~\cite{Baltieri_Vezzani_Cucchiara:2011}, VIPeR~\cite{Gray_Tao:2008}, CUHK01~\cite{Li_Wang:2013}, CUHK03~\cite{Li_Zhao_Xiao:2014}, Market1501~\cite{Zheng_Shen_Tian:2015} and DukeMTMCre-ID~\cite{Ristani_Solera_Zou:2016}, which are briefly introduced in the following paragraphs.~\footnote{The 3DPeS dataset provides the foreground masks, and the foreground masks of images in other datasets are obtained by using the algorithm~\cite{Zheng_Jayasumana_Romera:2015} in link \scriptsize{\url{http://www.robots.ox.ac.uk/~szheng/CRFasRNN.html}}.} The 3DPeS dataset contains 1,011 images of 192 persons captured from 8 outdoor cameras with different viewpoints, and each person has 2 to 26 images. The VIPeR dataset contains 632 person images captured by two cameras in an outdoor environment, and each person has only one image in each camera view. The CUHK01 dataset contains 971 persons captured from two camera views in a campus environment, and there are two images for each person under every camera view. The CUHK03 dataset contains 14,097 images from 1,467 persons, which is captured from six cameras in a campus environment and each person only has two camera views. The Market1501 dataset contains 32,668 images of 1,501 persons in a campus environment, in which each person is captured by six cameras at most, and two cameras at least. The DukeMTMC-reID dataset is consisted of 1,812 identities captured from 8 different cameras, in which 16,522 samples from 702 identities are used for training, 2,228 samples of another 702 identities are used as queries, and the remaining 17,661 samples are used for the gallery set.

{\bf Parameter settings.} The parameters are taken as follows: The weights are initialized from two zero-mean Gaussian distributions with the standard deviations of $0.01$ to $0.001$, and the bias terms are set as $0$. The learning rate $\omega = 0.01$, and decayed
by 0.1 at every 10,000 iterations, the margin parameter $\mathrm{M}=0.1$, the kernel parameters $\rho = 3, \sigma = 0.01$, the weight parameters $\zeta = 0.02$ and $\eta = 0.05$, the initial adaptive weights $u = 0.6$ and $v = 0.4$, and the weight updating rate $\gamma = 0.01$. If not specified, we use the same parameters in all the experiments.

\begin{table*}[t]
	\caption{The matching rates(\%) comparison with the state-of-the-art methods on the CUHK01 and CUHK03 datasets, in which `-' means they do not report the corresponding result.}
	\vspace{-0.1cm}
	\begin{center}
		\label{tab_3}
		\begin{tabular}{ |p{3.0cm}<{\centering} | c | c | c | c || c |c | c || c | c | c || c | c | c |}
			\hline
			\multicolumn{1}{|c|}{\multirow{2}{*}{Methods}} &
			\multicolumn{1}{c|}{\multirow{2}{*}{Year}} &
			\multicolumn{3}{c||}{CUHK01~{\tiny(p=100)}} &
			\multicolumn{3}{c||}{CUHK01~{\tiny(p=486)}} &
			\multicolumn{3}{c||}{CUHK03~{\tiny(p=100)}}  &
			\multicolumn{3}{c|}{CUHK03~{\tiny(p=700)}}\\
			\cline{3-14}
			&&Top 1 & Top 5 & Top10 &Top 1 & Top 5 & Top10 &Top 1 & Top 5 & Top10 &Top 1 & Top 5 & mAP\\
			\hline
			\hline
			kLFDA~\cite{Xiong_Gou_Camps:2014}       & 2014 & 42.7 & 69.0 & 79.6 & 32.7 & 59.0 & 69.6 & 48.2 & 59.3 & 66.4 &-&-&-\\
			LOMO+XQDA~\cite{Liao_Hu_Zhu:2015}       & 2015 & 77.6 & 94.1 & 97.5 & 63.2 & 83.9 & 90.0 & 52.0 & 82.2 & 92.1 & 14.8 &  -  & 13.6\\
			IDLA~\cite{Ahmed_Jones_Marks:2015}      & 2015 & 65.0 & 89.5 & 93.0 & 47.5 & 71.5 & 80.0 & 54.7 & 86.5 & 94.0 &-&-& -\\
			ITML~\cite{Davis_Kulis_Jain:2007}       & 2017 & 17.1 & 42.3 & 55.1 & 16.0 & 35.2 & 45.6 & 5.5  & 18.9 & 30.0 &   -  &  -  &  -  \\
			SVDNet~\cite{Sun_Zheng_Deng:2017}       & 2017 & - & - & - & - & - & -  & - & 95.2 & 97.2 & 40.9 & - &37.8\\
			PAN~\cite{Zheng_Zheng_Yang:2017}        & 2017 & - & - & - & - & - & -  & - &  -  &  -  & 36.9 & 56.9 & 35.0\\
			Quadruplet~\cite{Chen_Chen_Zhang:2017}  & 2017 & 79.0 & 96.0 & 97.0 & 62.6 & 83.0 & 88.8 & 74.5 & 96.6 & 99.0 & - & - & -\\
			DPFL~\cite{Chen_Zhu_Gong:2017}          & 2017 & - & - & - & - & - & - & - & - &  - & 43.0 & - & 40.5\\
			MLFN~\cite{Chang_Hospedales_Xiang:2018} & 2018 & - & - & - & - & - & - & 82.8 & - & - & 54.7 & - & 49.2\\
			PRGP~\cite{Tian_Yi_Li:2018}             & 2018 & - & - & - & 80.7 & 95.0 & 97.5 & 91.7 & 98.2 & 98.7 & - & - & -\\
			MLS~\cite{Guo_Cheung:2018}              & 2018 & 88.2 & 98.2 & 99.4 & - & - & - & 87.5 & 97.9 & 99.5 & - & - & -\\
			HA-CAN~\cite{Li_Zhu_Gong:2018}          & 2018 & - & - & - & - & - & - & - & - & - & 44.4 & - & 41.0\\
			DGRW~\cite{Shen_Li_Xiao:2018}           & 2018 & - & - & - & - & - & - & \bf{94.9} & 98.7 & 99.3 & - & - & -\\
			MGCAN~\cite{Song_Huang_Ouyang:2018}     & 2018 & - & - & - & - & - & - & - & - & - & 50.1 & - & 50.2\\
			BraidNet~\cite{Wang_Chen_Wu:2018}       & 2018 & 93.0 & - & 99.9 & - & - & - & 88.2 & - & 98.7 & - & - & -\\
			AACN~\cite{Xu_Zhao_Zhu:2018}            & 2018 & 88.1 & 96.7 & 98.2 & - & - & - & 91.4 & 98.9 & 99.5 & - & - & -\\
			PN-GAN~\cite{Qian_Fu_Xiang:2018}        & 2018 & - & - & - & 67.7 & 86.6 & 91.8 & 79.8 & 96.2 & 98.6 & - & - & -\\
			DaRe~\cite{Wang_Wang_You:2018}          & 2018 & - & - & - & - & - & - & - & - & - & 66.1 & - & 66.7 \\
			\hline
			Our FANN & 2018 &\bf{98.1} & \bf{99.8} & \bf{100} & \bf{81.2} & \bf{95.3} & \bf{99.1} & 92.3 & \bf{99.2} & \bf{100} &\bf{70.2} &  \bf{86.1} & \bf{70.4}\\
			\hline
		\end{tabular}
	\end{center}
	\vspace{-0.3cm}
\end{table*}

\begin{table}[t]
	\caption{The matching rates(\%) comparison with the state-of-the-art methods on the 3DPeS dataset, in which `-' means they do not report the corresponding result.}
	\vspace{-0.1cm}
	\begin{center}
		\label{tab_1}
		\begin{tabular}{|p{1.7cm}<{\centering} | p{0.5cm}<{\centering}| p{0.7cm}<{\centering} |  p{0.7cm}<{\centering} |  p{0.7cm}<{\centering}|  p{0.7cm}<{\centering} |  p{0.7cm}<{\centering}|}
			\hline
			Methods & Year & Top 1 & Top 5 & Top10 & Top15 & Top20\\
			\hline
			\hline
		    KISSME~\cite{Koestinger_Hirzer_Wohlhart:2012} & 2012 & 22.9 & 48.7 & 62.2 & 72.4 & 78.1\\
			LF~\cite{Pedagadi_Orwell_Velastin:2013}       & 2013 & 33.4 & 45.5 & 69.9 & 76.5 & 81.0\\
			kLFDA~\cite{Xiong_Gou_Camps:2014}             & 2014 & 54.0 & 77.7 & 85.9 & 90.0 & 92.4\\
			MFA~\cite{Xiong_Gou_Camps:2014}               & 2014 & 41.8 & 65.5 & 75.7 &   -  & 85.2\\
			ME~\cite{Paisitkriangkrai_Shen_Van:2015}      & 2015 & 53.3 & 76.8 & 86.0 & 89.4 & 92.8\\
			SCSP~\cite{Chen_Yuan_Chen:2016}               & 2016 & 57.3 & 78.9 & 85.0 & 89.5 & 91.5\\
			JSTL~\cite{Xiao_Li_Ouyang:2016}               & 2016 & 56.0 &   -  &   -  &   -  &   - \\
		    WARCA~\cite{Jose_Fleuret:2016}                & 2016 & 51.9 & 75.6 &   -  &   -  &   - \\
			Spindle~\cite{Zhao_Tian_Sun:2017}             & 2017 & 62.1 & 83.4 & 90.5 &   -  & 95.7\\
			P2S~\cite{Zhou_Wang_Wang:2017}                & 2017 & 71.2 & 90.5 & 95.2 & 96.9 & 97.6\\
			SPL~\cite{Zhou_Wang_Meng:2017}                & 2018 & 72.2 & 90.7 & 95.3 & 96.8 & 97.5\\
			PRGP~\cite{Tian_Yi_Li:2018}                   & 2018 & 64.1 & 87.4 & 90.4 &   -  & 93.7\\
			\hline
			Our FANN                             & 2018 &\bf{78.9}&\bf{92.3}&\bf{95.7}&\bf{98.1}&\bf{99.4}  \\
			\hline
		\end{tabular}
	\end{center}
    \vspace{-0.3cm}
\end{table}

\begin{table}[t]
	\caption{The matching rates(\%) comparison with the state-of-the-art methods on the VIPeR dataset, in which `-' means they do not report the corresponding result.}
	\vspace{-0.1cm}
	\begin{center}
		\label{tab_2}
		\begin{tabular}{|p{1.7cm}<{\centering}| p{0.5cm}<{\centering}| p{0.7cm}<{\centering} |  p{0.7cm}<{\centering} |  p{0.7cm}<{\centering}|  p{0.7cm}<{\centering} |  p{0.7cm}<{\centering}|}
			\hline
			Methods & Year & Top 1 & Top 5 & Top10 & Top15 & Top20\\
			\hline
			\hline
			RPLM~\cite{Hirzer_Roth:2012}                   & 2012 & 27.3 & 55.3 & 69.0 & 77.1 & 82.7\\
			sLDFV~\cite{Ma_Su_Jurie:2012}                  & 2012 & 26.5 & 56.4 & 70.9 &   -  & 84.6\\
			kBiCov~\cite{Ma_Su_Jurie:2014}                 & 2015 & 31.1 & 58.3 & 70.7 &   -  & 82.4\\
			Triplet~\cite{Ding_Lin_Wang:2015}              & 2015 & 40.5 & 60.8 & 70.4 & 78.4 & 84.4\\
			LNDS~\cite{Zhang_Xiang_Gong:2016}              & 2016 & 51.2 & 82.1 & 90.5 &   -  &   - \\
			Quadruplet~\cite{Chen_Chen_Zhang:2017}         & 2017 & 49.1 & 73.1 & 81.9 &   -  &   - \\
			Spindle~\cite{Zhao_Tian_Sun:2017}              & 2017 & 53.8 & 74.1 & 83.2 &   -  & 92.1\\
			SSM~\cite{Bai_Bai_Tian:2017}                   & 2017 & 53.7 &   -  &   -  &   -  & 96.1\\
			PDC~\cite{Su_Li_Zhang:2017}                    & 2017 & 51.3 & 74.0 & 84.2 &   -  & 91.5\\
			SPL~\cite{Zhou_Wang_Meng:2017}                 & 2018 & 56.3 & 83.0 & 92.0 & 93.8 & 95.9\\
			PRGP~\cite{Tian_Yi_Li:2018}                    & 2018 & 50.6 & 70.3 & 79.1 &   -  & 88.0\\
			MLS~\cite{Guo_Cheung:2018}                     & 2018 & 50.0 & 73.1 & 84.4 &   -  &   - \\
			\hline
			Our FANN                                       & 2018 &\bf{58.4} &\bf{83.7} &\bf{92.2} &\bf{93.9} &\bf{96.4}  \\
			\hline
		\end{tabular}
	\end{center}
	\vspace{-0.3cm}
\end{table}

{\bf Evaluation protocol.} Our experiments use the Cumulative Matching Characteristic~(CMC) curve to measure the performance, which is an estimation of finding the corrected top $n$ match. For the 3DPeS and VIPeR datasets, we follow the single-shot protocol in~\cite{Ding_Lin_Wang:2015}, in which 96 persons from the 3DPeS dataset and 316 persons in the VIPeR dataset are randomly chosen to train the deep neural network, and the remaining identities are used to evaluate the performance. For the CUHK01 and CUHK03 datasets, we follow two data partition protocols to split the datasets into the training sets and testing sets. Specifically, 100/486 persons of the CUHK01 dataset and 100/700 persons of the CUHK03 dataset are used to evaluate the performance, and the remainings are used to train the deep neural network. For the Market1501 and DukeMTMC-reID datasets, we used the provided data partition methods to prepare the training and testing samples. Besides, the mean Average Precision~(mAP) is also used to evaluate the performance on the CUHK03, Market1501 and DukeMTMC-reID datasets. To obtain a statistical result, we repeated the testing 10 times to report the average result.

\subsection{Comparison Results}
Firstly, we will compare our method with the state-of-the-art approaches on the six public benchmark datasets, respectively. Secondly, the performances of attention learning based methods will be solely evaluated, so as to compare how much they can improve the final results. For clarity, we highlight the best results in bold.

{\bf Comparisons with the state-of-the-arts.} The detailed results are shown in Table~\ref{tab_1} to Table~\ref{tab_4}, from which we can see that our FANN has achieved the competitive results on nearly all of the six public benchmark datasets. Specifically, our FANN outperforms the previous best performed SPL~\cite{Zhou_Wang_Meng:2017} method by $6.7\%$ on the 3DPeS dataset in the Top 1 accuracy. Besides, our FANN also outperforms the previous best performed SPL~\cite{Zhou_Wang_Meng:2017} method by $2.1\%$ on the VIPeR dataset in the Top 1 accuracy. For the CUHK01 and CUHK03 datasets, our FANN outperforms the previous best performed BraidNet~\cite{Wang_Chen_Wu:2018} 
by $5.1\%$, while lags behind the previous best performed DGRW~\cite{Shen_Li_Xiao:2018} method by $2.6\%$ in the Top 1 accuracy, when $100$ identities are randomly chosen to evaluate the performance, respectively. When $486$ identities from the CUHK01 dataset are used to evaluate the performance, our FANN outperforms the previous best performed PRGP~\cite{Tian_Yi_Li:2018}  method by $0.5\%$ in the Top 1 accuracy. In addition, our FANN outperforms the previous best performed DaRe~\cite{Wang_Wang_You:2018} by $4.1\%$ and $3.7\%$ in terms of the Top 1 accuracy and mAP, when $700$ identities are used to evaluate the performance on the CUHK03 dataset, respectively. The same conclusion can be get on the Market1501 and DukeMTMC-reID datasets using the single-query evaluation, in which our FANN  outperforms the previous best performed GCSL~\cite{Chen_Xu_Li:2018} and PCB~\cite{Sun_Zheng_Yang:2018} methods by $0.6\%, 0.3\%$ and $0.9\%, 0.7\%$ in the Top 1 accuracy and mAP on the Market1501 and DukeMTMC-reID datasets, respectively.

\begin{table*}[t]
	\caption{The matching rates(\%) improved by each of our contributions on the six benchmark datasets, respectively.}
	\vspace{-0.1cm}
	\begin{center}
		\label{tab_5}
		\begin{tabular}{| c | c | c | c | c | c | c | c | c | c | c | c | c | c | c | c | c |}
			\hline
			\multicolumn{1}{|c|}{\multirow{2}{*}{Methods}} &
			\multicolumn{2}{c|}{3DPeS} &
			\multicolumn{2}{c|}{VIPeR} &
			\multicolumn{2}{c|}{CUHK01~{\tiny(100)}}  &
			\multicolumn{2}{c|}{CUHK01~{\tiny(486)}}  &
			\multicolumn{2}{c|}{CUHK03~{\tiny(100)}}  &
			\multicolumn{2}{c|}{CUHK03~{\tiny(700)}}  &
			\multicolumn{2}{c|}{Market.}  &
			\multicolumn{2}{c|}{Duke.}\\
			\cline{2-17}
			& \hspace{-0.15cm} Top 1 \hspace{-0.15cm} & \hspace{-0.15cm} Top 5 \hspace{-0.15cm} & \hspace{-0.15cm} Top 1 \hspace{-0.15cm}
			& \hspace{-0.15cm} Top 5 \hspace{-0.15cm} & \hspace{-0.15cm} Top 1 \hspace{-0.15cm} & \hspace{-0.15cm} Top 5 \hspace{-0.15cm}
			& \hspace{-0.15cm} Top 1 \hspace{-0.15cm} & \hspace{-0.15cm} Top 5 \hspace{-0.15cm} & \hspace{-0.15cm} Top 1 \hspace{-0.15cm}
			& \hspace{-0.15cm} Top 5 \hspace{-0.15cm} & \hspace{-0.15cm} Top 1 \hspace{-0.15cm} & \hspace{-0.15cm} mAP \hspace{-0.15cm}
			& \hspace{-0.15cm} Top 1 \hspace{-0.15cm} & \hspace{-0.15cm} mAP \hspace{-0.15cm} & \hspace{-0.15cm} Top 1 \hspace{-0.15cm}
			& \hspace{-0.15cm} mAP \hspace{-0.15cm}\\
			\hline
			\hline
			\textbf{Baseline1}                    & 67.3 & 89.2 & 47.3 & 73.1 & 79.8 & 90.4 & 64.1 & 86.1 & 73.9 & 92.1 & 52.6 & 51.8 & 67.6 & 45.4 & 64.4 & 43.1 \\
			\textbf{Baseline2}                    & 65.4 & 87.1 & 43.2 & 72.2 & 75.8 & 86.2 & 58.5 & 82.2 & 68.9 & 89.1 & 48.4 & 47.9 & 62.2 & 39.6 & 60.1 & 39.6 \\
			\textbf{S}                            & 72.5 & 90.7 & 50.9 & 80.8 & 92.1 & 95.9 & 72.3 & 89.6 & 81.4 & 95.2 & 62.1 & 62.3 & 78.4 & 54.1 & 72.1 & 60.1\\
			\textbf{L}                            & 73.1 & 90.9 & 51.2 & 81.1 & 92.7 & 96.8 & 74.4 & 91.2 & 83.6 & 96.4 & 65.0 & 63.7 & 84.6 & 64.7 & 77.6 & 64.2\\
			\textbf{F}                            & 72.1 & 89.7 & 50.1 & 80.2 & 90.1 & 95.3 & 74.1 & 90.8 & 82.4 & 96.1 & 61.9 & 60.7 & 77.9 & 60.9 & 71.4 & 58.4\\
			\textbf{S} + \textbf{F}                 & 75.2 & 91.8 & 55.1 & 81.9 & 94.2 & 98.1 & 76.5 & 92.5 & 88.1 & 97.1 & 66.7 & 65.9 & 86.1 & 67.6 & 77.1 & 63.6\\
			\textbf{S} + \textbf{L}                 & \textbf{78.9} & \textbf{92.3} & \textbf{58.4} & \textbf{83.7} & \textbf{98.1} & \textbf{99.8} & \textbf{81.2} & \textbf{95.3} & \textbf{92.3} & \textbf{99.2} & \textbf{70.2} & \textbf{69.5} & \textbf{94.4} & \textbf{82.5} & \textbf{85.2} & \textbf{70.2}\\
			\hline
		\end{tabular}
	\end{center}
	\vspace{-0.3cm}
\end{table*}

\begin{table}[t]
	\caption{The matching rates(\%) comparison with the state-of-the-art methods on the Market1501 and DukeMTMC-reID datasets, in which `-' means they do not report the corresponding result.}
	\vspace{-0.1cm}
	\begin{center}
		\label{tab_4}
		\begin{tabular}{ |c| c | p{0.8cm}<{\centering} | p{0.8cm}<{\centering} | p{0.8cm}<{\centering} | p{0.8cm}<{\centering}|}
			\hline
			\multicolumn{1}{|c|}{\multirow{2}{*}{Methods}} &
			\multicolumn{1}{c|}{\multirow{2}{*}{Year}} &
			\multicolumn{2}{c|}{Market.} &
			\multicolumn{2}{c|}{Duke.}\\
			\cline{3-6}
			&& Top 1  &  mAP  & Top 1 & mAP \\
			\hline
			LDNS~\cite{Zhang_Xiang_Gong:2016}                    & 2016 & 61.0 & 35.6 &  -   &  -\\
			S2S~\cite{Zhou_Wang_Shi:2018}                        & 2017 & 65.3 & 40.0 & - & - \\
			DSPL~\cite{Zhou_Wang_Meng:2017}                      & 2017 & 72.9 & 46.7 &   -  &   - \\
			JLML~\cite{Li_Zhu_Gong:2017}                         & 2017 & 83.9 & 64.4 &  -   &   - \\
			PDC~\cite{Su_Li_Zhang:2017}                          & 2017 & 84.1 & 63.4 &  -   &   - \\
			SVDNet~\cite{Sun_Zheng_Deng:2017}                    & 2017 & 82.3 & 62.1 & 76.7 & 56.8\\
			SSM~\cite{Bai_Bai_Tian:2017}                         & 2017 & 82.2 & 68.8 &  -   &  -\\
			DPFL~\cite{Chen_Zhu_Gong:2017}                       & 2017 & 88.6 & 72.6 & 79.2 & 60.6\\
			PRGP~\cite{Tian_Yi_Li:2018}                          & 2017 & 81.2 & - & - & - \\
			MLFN~\cite{Chang_Hospedales_Xiang:2018}              & 2018 & 90.0 & 74.3 & 81.0 & 62.8\\
			DGRW~\cite{Shen_Li_Xiao:2018}                        & 2018 & 92.7 & 82.5 & 80.7 & 66.4\\
			DuATM~\cite{Si_Zhang_Li:2018}                        & 2018 & 91.4 & 76.6 & 81.8 & 64.6\\
			BraidNet~\cite{Wang_Chen_Wu:2018}                    & 2018 & 83.7 & 69.5 & 76.4 & 59.5\\
			AACN~\cite{Xu_Zhao_Zhu:2018}                         & 2018 & 85.9 & 66.9 & 76.8 & 59.3\\
			SGGNN~\cite{Shen_Li_Yi:2018}                         & 2018 & 92.3 & 82.8 & 81.8 & 68.2\\
			PN-GAN~\cite{Qian_Fu_Xiang:2018}                     & 2018 & 89.4 & 72.6 & 73.6 & 53.2\\
			GCSL~\cite{Chen_Xu_Li:2018}                          & 2018 & 93.5 & 81.6 & 84.9 & 69.5\\
			PCB~\cite{Sun_Zheng_Yang:2018}                       & 2018 & 93.8 & 81.6 & 83.3 & 69.2\\
			\hline
			Our Method (FANN)                                    & 2018 & \bf{94.4}   & \bf{82.5} & \bf{85.2}  &\bf{70.2}\\
			\hline
		\end{tabular}
	\end{center}
	\vspace{-0.3cm}
\end{table}

\begin{figure}[t]
	\centering
	\begin{tabular}{c}
		\includegraphics[height = 4.0cm, width = 7.5cm]{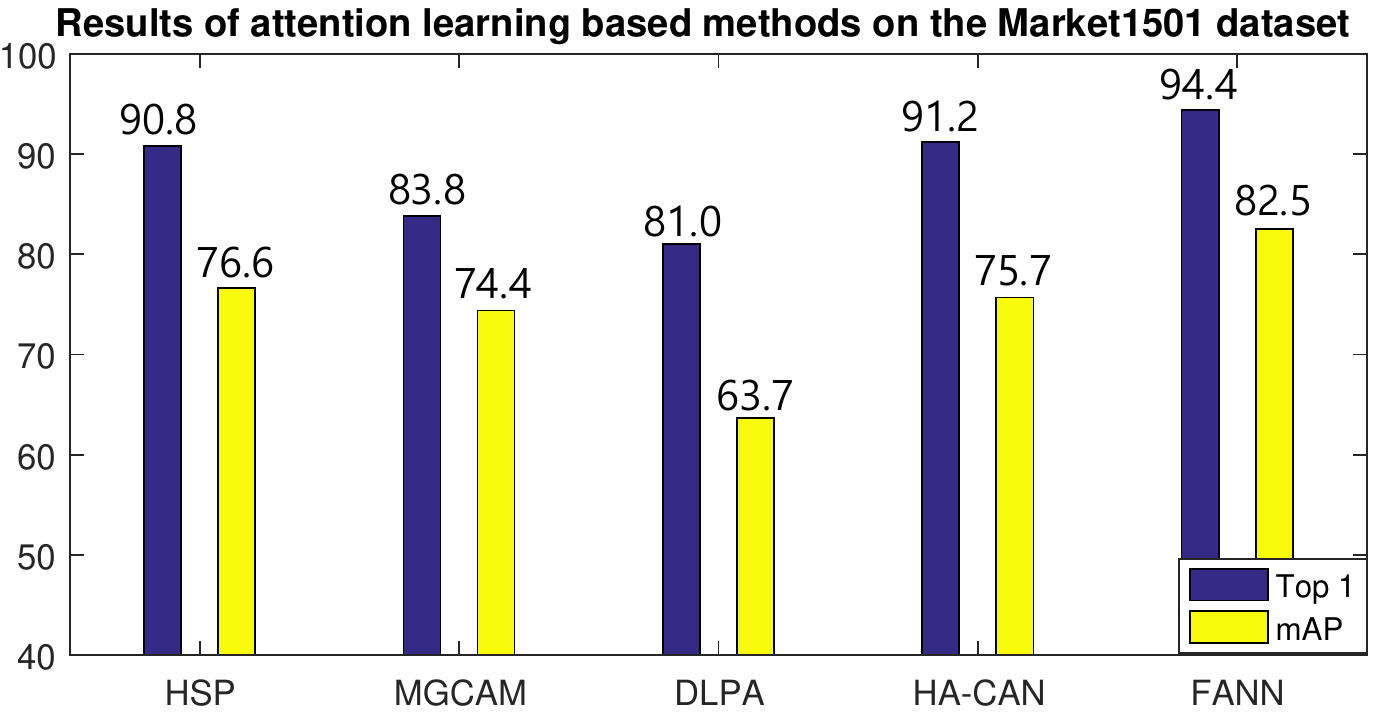}
	\end{tabular}
	\vspace{-0.1cm}
	\caption{Comparison between our method and the other attention learning based approaches on the Market1501 dataset, in which the HSP~\cite{Kalayeh_Basaran_Gkmen:2018}, MGCAM~\cite{Song_Huang_Ouyang:2018} and our FANN belong to the supervised approaches, and the DLPA\cite{Zhao_Li_Zhuang:2017}  and HA-CAN~\cite{Li_Zhu_Gong:2018} are the unsupervised methods.}
	\label{fig_9}
	\vspace{-0.3cm}
\end{figure}

{\bf Comparisons of attention learning.} The attention learning based methods can usually improve the discriminative ability of learned features in solving the person Re-ID problem, because they can further address the foreground persons in the training process. As discussed above, the supervised methods often outperform the unsupervised ones in the final accuracy. In Fig~\ref{fig_9}, we compare our method with the other four attention learning based methods on the Market1501 dataset, in which the HSP~\cite{Kalayeh_Basaran_Gkmen:2018} and MGCAM~\cite{Song_Huang_Ouyang:2018} are the supervised methods, while the DLPA\cite{Zhao_Li_Zhuang:2017} and HA-CAN~\cite{Li_Zhu_Gong:2018} are the unsupervised methods. From the results, we can see that the worse results are obtained by the DLPA, and the best performances are achieved by our FANN. Besides, we also notice that the HA-CAN significantly outperforms the MGCAM in the Top 1 accuracy, which indicates that it is possible to learn the attentive features in an unsupervised manner. In the future study, we will strive to design an attention mechanism in network, so as to improve the feature representation capability without using the expensive foreground annotations.

\subsection{Ablation Study}
Firstly, we will evaluate how much each of our contributions improves the final person Re-ID results. Secondly, the effectiveness of our FANN in background suppression will be illustrated, including the visualization of learned feature maps and the robustness of our FANN to different ground-truth masks. Then, we will show the robustness of our method to different parameter settings and study how to set the number of residual blocks in the body part subnetwork. Finally, some ranking examples will be presented and discussed.

{\bf Improvements by each contribution.} In order to show how much each contribution improves the final results, we carefully design seven different experiments on each dataset, as shown in Table~\ref{tab_5}. In particular, \textbf{Baseline 1} denotes the performances that we get rid of the decoder network and takes the asymmetric triplet loss function to train the remaining network, and \textbf{Baseline 2} indicates the results that we use the masked foreground images to replace the inputs in \textbf{Baseline 1}. Besides, \textbf{S} means the performances that we get rid of the decoder network and takes our symmetric triplet loss function to train the remaining network, \textbf{L} represents the results that we use the asymmetric triplet loss function and local regression loss function to train the whole network, and \textbf{F} indicates the performances that we use the asymmetric triplet loss function and Euclidean loss function to train the whole network. What's more, \textbf{S} + \textbf{F} denotes the results that we use our symmetric triplet loss function and Euclidean loss function to train the whole network, and \textbf{S} + \textbf{L} represents the results obtained by jointly using the symmetric triplet loss function and local regression loss function, which is actually equivalent to our FANN method.

From the results we can see that the \textbf{S} + \textbf{L} significantly outperforms the other six situations on all the six benchmark datasets, which can well explain the effectiveness of our symmetric triplet loss function, the local regression loss function and the neural network in learning the discriminative features from the foregrounds of input images. For simplicity, we take the results on the VIPeR dataset to explain the detailed improvements: 1) Compare the performances in \textbf{Baseline 1} and \textbf{Baseline 2}, we can find that directly feeding the masked images to train the neural network can not improve the person Re-ID results, because the strong edge responses brought by the masks are harmful to learn the discriminative features. 2) Compare the results between \textbf{Baseline 1} and \textbf{S}, between \textbf{F} and \textbf{S + F}, between \textbf{L} and \textbf{S + L}, we can find that our symmetric triplet loss function can improve the Top 1 accuracy by $5.2\%$, $3.1\%$, $5.8\%$ in the three cases, respectively. 3) For the improvements of our local regression loss function, we compare the results between \textbf{Baseline 1} and \textbf{L}, between \textbf{F} and \textbf{L}, between \textbf{S + F} and \textbf{S + L}, and the results show that our local regression loss function can improve the Top 1 accuracy by $5.8\%$, $1.0\%$, $3.7\%$ in the three cases, respectively. 4) Finally, we evaluate the improvements bought by our neural network by comparing the results between \textbf{Baseline 1} and \textbf{F},  between \textbf{Baseline 1} and \textbf{L}, between \textbf{S} and \textbf{S + F}, which show that our FANN improves the Top 1 accuracy by $4.8\%$, $5.8\%$, $2.7\%$ in the three cases, respectively. What's more, the same conclusions can be found if we evaluate the performances on the other five datasets.

\begin{figure}[t]
	\centering
	\begin{tabular}{c}
		\includegraphics[height = 5.2cm, width = 7.0cm]{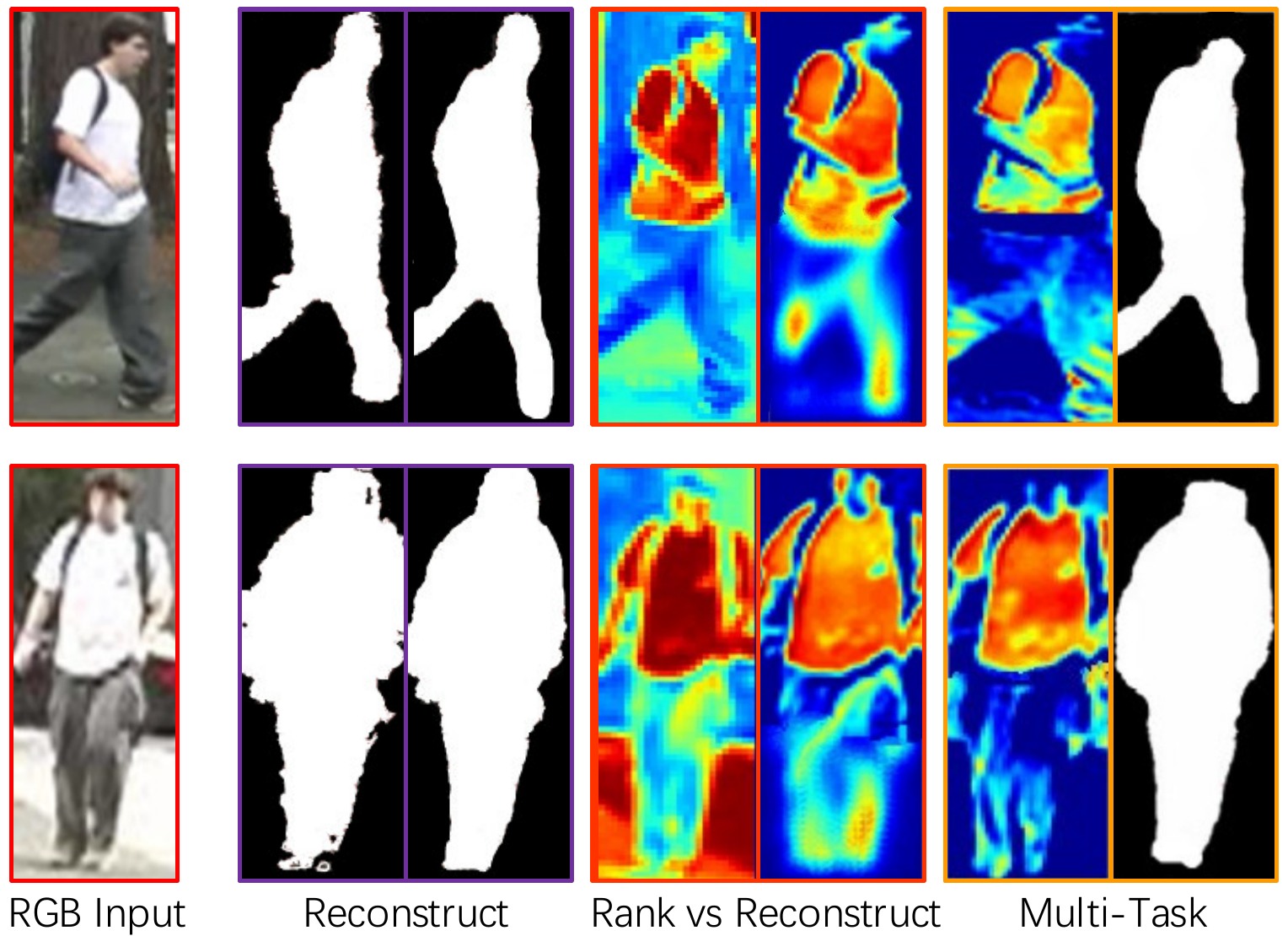}
	\end{tabular}
	\vspace{-0.1cm}
	\caption{Illustration of the learned feature maps and masks by our neural network. From left to right: the RGB inputs, the ground truth masks by using the off-the-shelf segmentation method, the foreground masks obtained in the reconstruction task, the feature maps learned in the ranking and reconstruction tasks, the feature maps and foreground masks obtained in the multi-task.}
	\label{fig_5}
	\vspace{-0.3cm}
\end{figure}

{\bf Effectiveness in background suppression.} In this paragraph, we will explain the internal reason of how our FANN focuses its attention on the foregrounds of input images. In our FANN, we apply the encoder-decoder mechanism to drive the attention, in which we reconstruct the mask of foreground at the output of decoder network, and the encoder network will be naturally regularized by the decoder network in the training process. As a result, the encoder network will pay more attention on the foregrounds of input images, which is effective to suppress noises in the background. Then, the resulting feature maps of encoder network will be fed into the subsequent networks for discriminative feature learning. Incorporating the mask reconstruction and feature learning into a multi-task learning framework, a discriminative feature representation can be learned to further improve the final person Re-ID results.

\begin{table}[t]
	\caption{The matching rate (\%) on our method by using different ground truth masks on the six benchmark datasets.}
	\begin{center}
		\label{tab_8}
		\begin{tabular}{| c | c | c | c | c | c | c |}
			\hline
			\multicolumn{1}{|c|}{\multirow{2}{*}{Datasets}} &
			\multicolumn{2}{c|}{Baseline} &
			\multicolumn{2}{c|}{Mask 1} &
			\multicolumn{2}{c|}{Mask 2}\\
			\cline{2-7}
			&Top 1 & Top 5 & Top 1  & Top 5 & Top 1 & Top 5\\
			\hline
			\hline
			3DPeS                            & 72.5 & 90.7 & 76.3 & 92.0 &\textbf{78.9}  & \textbf{92.3}\\
			VIPeR                            & 50.9 & 80.8 & 56.4 & 82.1 &\textbf{58.4}  &\textbf{83.7}\\
			CUHK01                           & 92.1 & 95.9 & 96.6 & 99.2 & \textbf{98.1} & \textbf{99.8}\\
			CUHK03                           & 81.4 & 95.2 & 91.3 & 98.7 & \textbf{92.3} & \textbf{99.2}\\
			Market1501                       & 78.4 & 54.1 & 92.1 & 94.9 & \textbf{94.4} & \textbf{95.3}\\
			DukeMTMC                         & 72.1 & 60.1 & 83.2 & 90.8 & \textbf{85.2} & \textbf{91.6}\\
			\hline
		\end{tabular}
	\end{center}
\end{table}

Some representative feature maps of two input images are shown in Fig.~\ref{fig_5}, in which the two images represent the same person under two disjoint camera views. Specifically, the second column shows the ground truth masks obtained by the segmentation method, and the third column represents the binary masks reconstructed by only running the reconstruction task. The results indicate that our local regression loss function is effective in reconstructing the binary mask. The fourth and fifth columns illustrate the feature maps of encoder network in the ranking and reconstruction tasks, which indicate that the reconstruction task can focus more attentions on the foreground than the ranking task. The sixth and seventh columns represent the feature maps of encoder network and the reconstructed masks by using the multi-task objective function, which illustrate that running the two tasks in an end-to-end manner is more beneficial to learn the foreground attentive features for person Re-ID.

\begin{figure*}[!htb]
	\centering
	\begin{tabular}{cccc}
		\hspace{-0.3cm}
		\includegraphics[height = 2.1cm, width = 4.5cm]{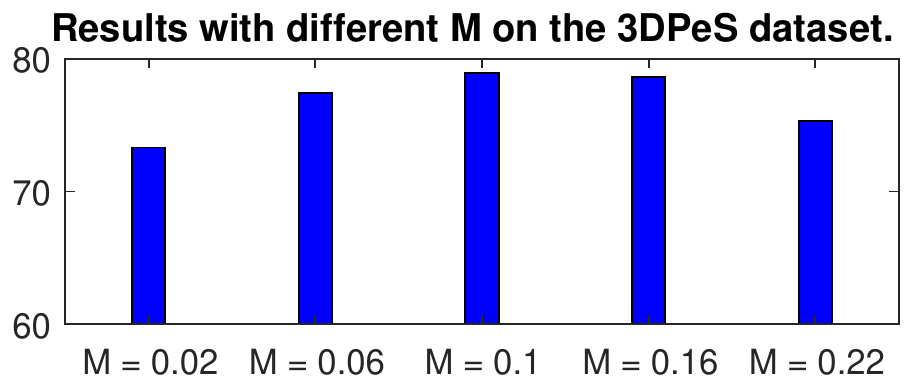}&
		\hspace{-0.5cm}
		\includegraphics[height = 2.1cm, width = 4.5cm]{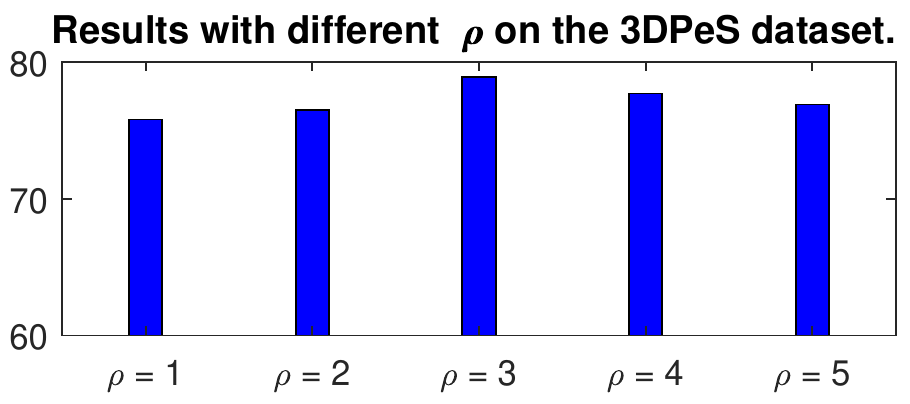}&
		\hspace{-0.5cm}
		\includegraphics[height = 2.1cm, width = 4.5cm]{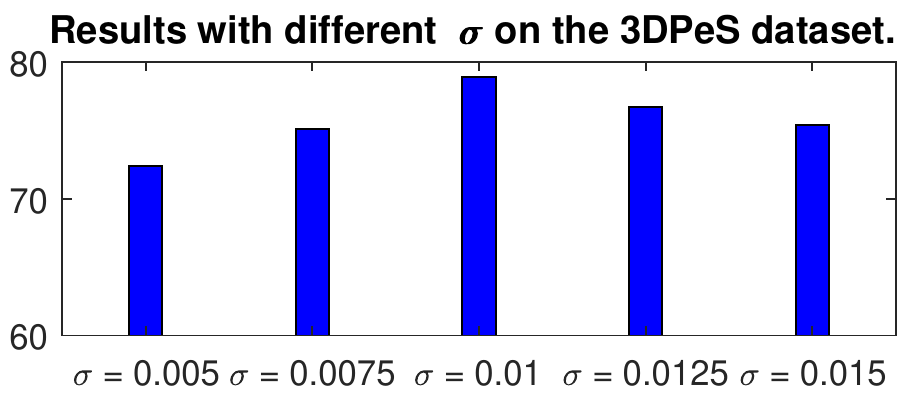}&
		\hspace{-0.5cm}
		\includegraphics[height = 2.1cm, width = 4.5cm]{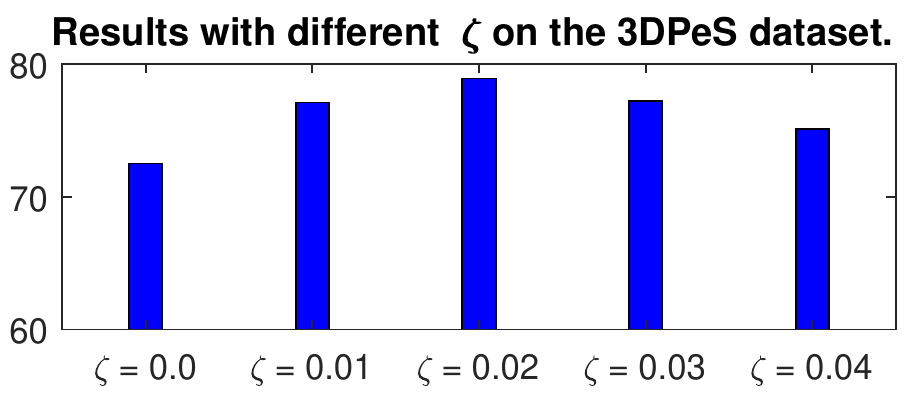}\\
		\hspace{-0.3cm}
		\includegraphics[height = 2.1cm, width = 4.5cm]{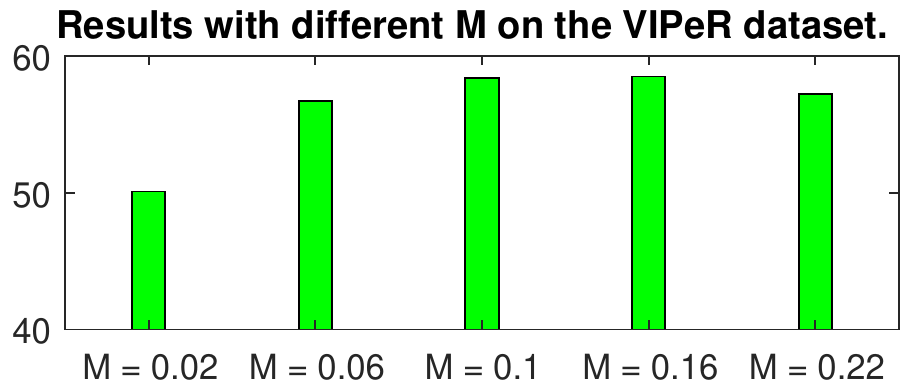}&
		\hspace{-0.5cm}
		\includegraphics[height = 2.1cm, width = 4.5cm]{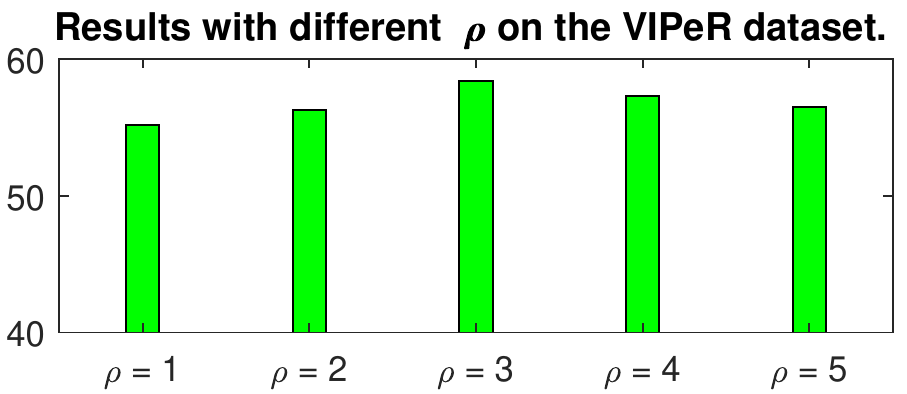}&
		\hspace{-0.5cm}
		\includegraphics[height = 2.1cm, width = 4.5cm]{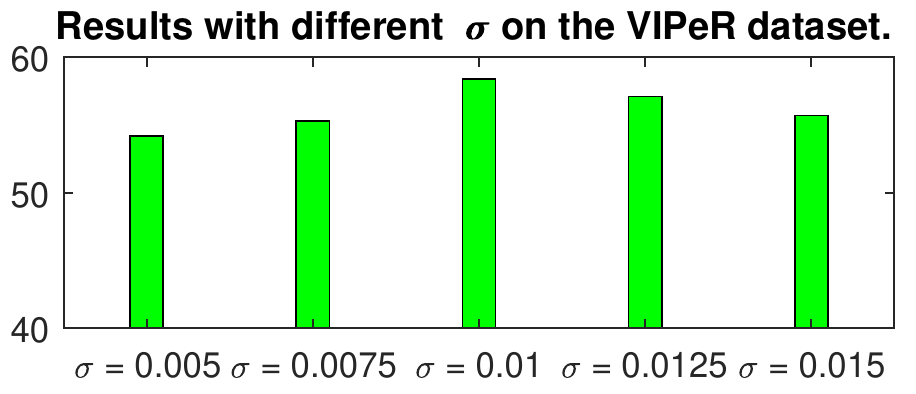}&
		\hspace{-0.5cm}
		\includegraphics[height = 2.1cm, width = 4.5cm]{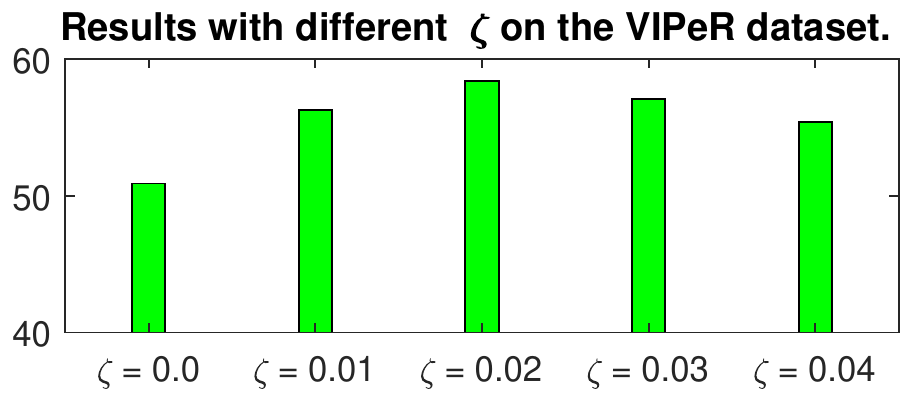}\\
		\hspace{-0.3cm}
		\includegraphics[height = 2.1cm, width = 4.5cm]{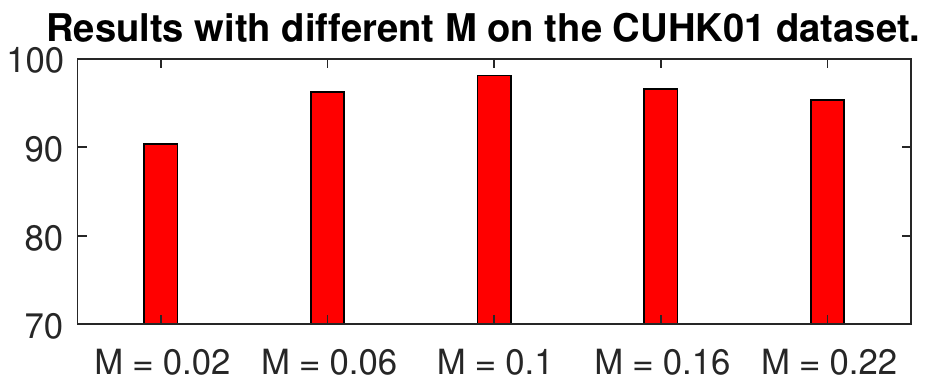}&
		\hspace{-0.5cm}
		\includegraphics[height = 2.1cm, width = 4.5cm]{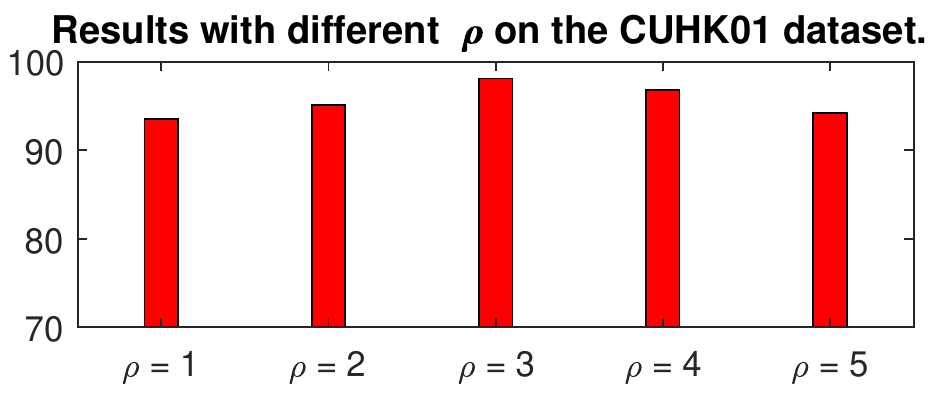}&
		\hspace{-0.5cm}
		\includegraphics[height = 2.1cm, width = 4.5cm]{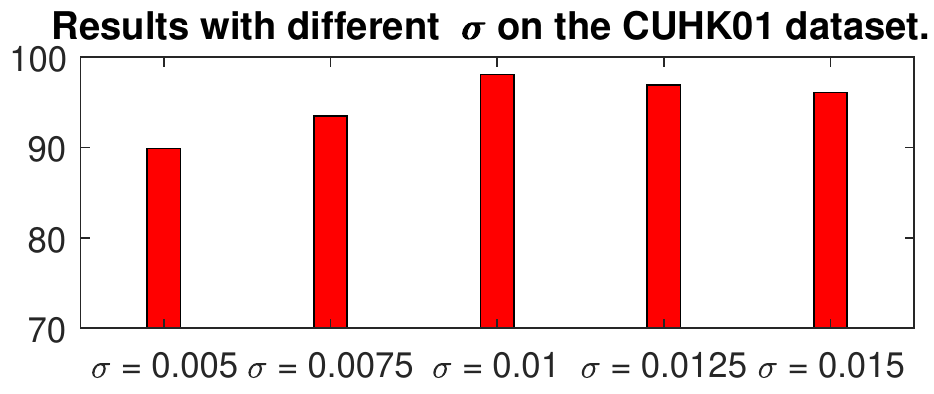}&
		\hspace{-0.5cm}
		\includegraphics[height = 2.1cm, width = 4.5cm]{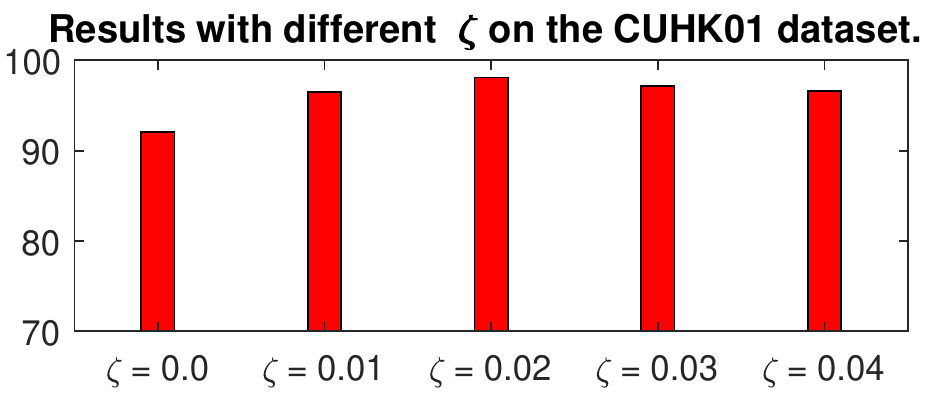}\\
		\hspace{-0.3cm}
		\includegraphics[height = 2.1cm, width = 4.5cm]{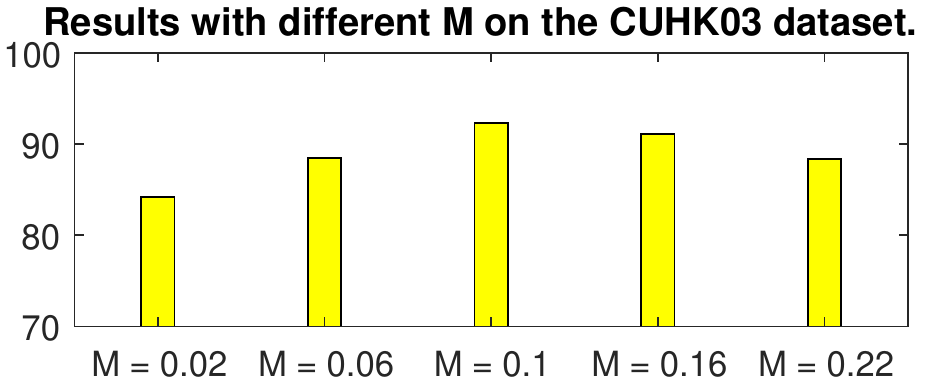}&
		\hspace{-0.5cm}
		\includegraphics[height = 2.1cm, width = 4.5cm]{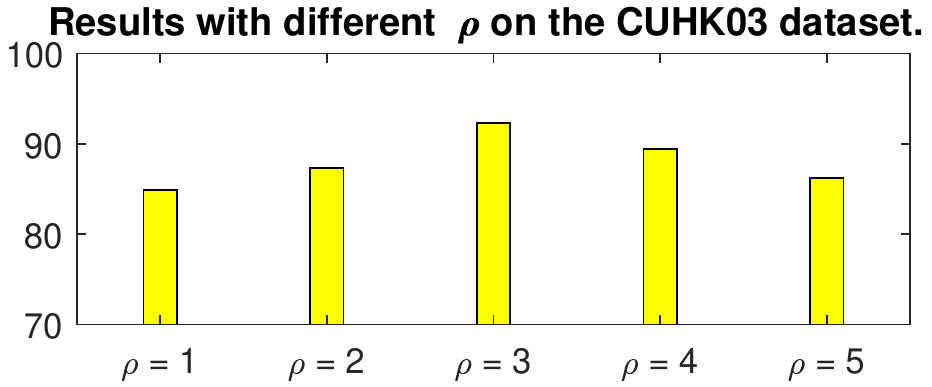}&
		\hspace{-0.5cm}
		\includegraphics[height = 2.1cm, width = 4.5cm]{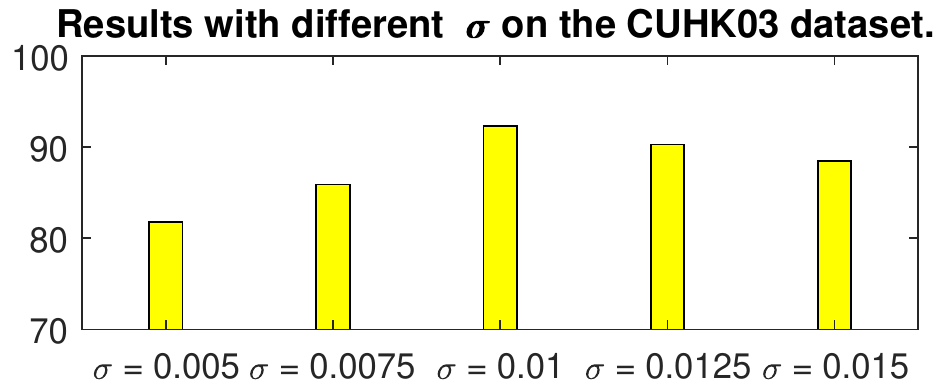}&
		\hspace{-0.5cm}
		\includegraphics[height = 2.1cm, width = 4.5cm]{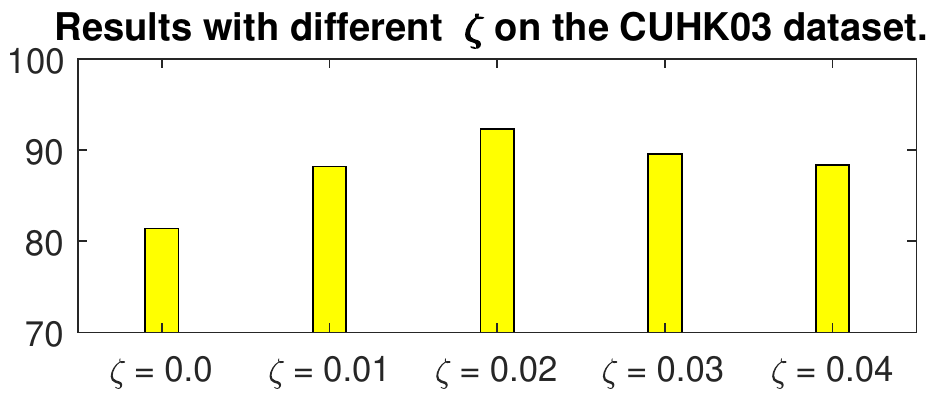}\\
		\hspace{-0.3cm}
		\includegraphics[height = 2.1cm, width = 4.5cm]{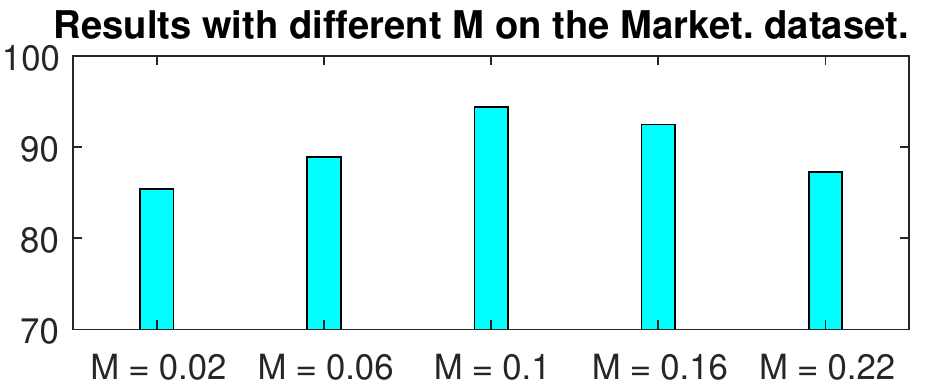}&
		\hspace{-0.5cm}
		\includegraphics[height = 2.1cm, width = 4.5cm]{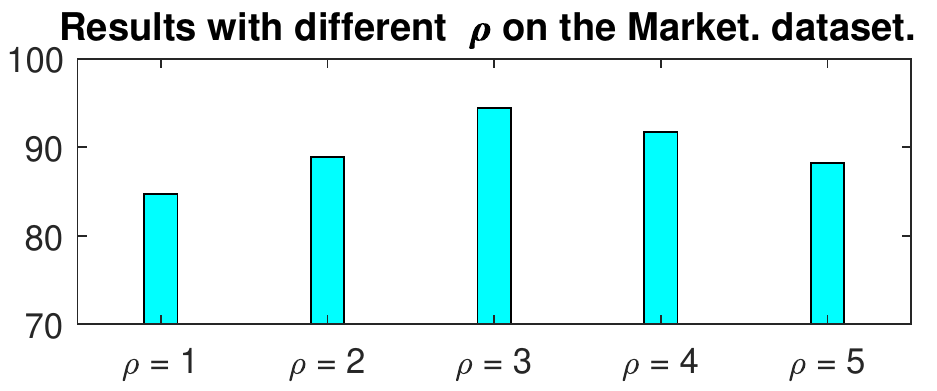}&
		\hspace{-0.5cm}
		\includegraphics[height = 2.1cm, width = 4.5cm]{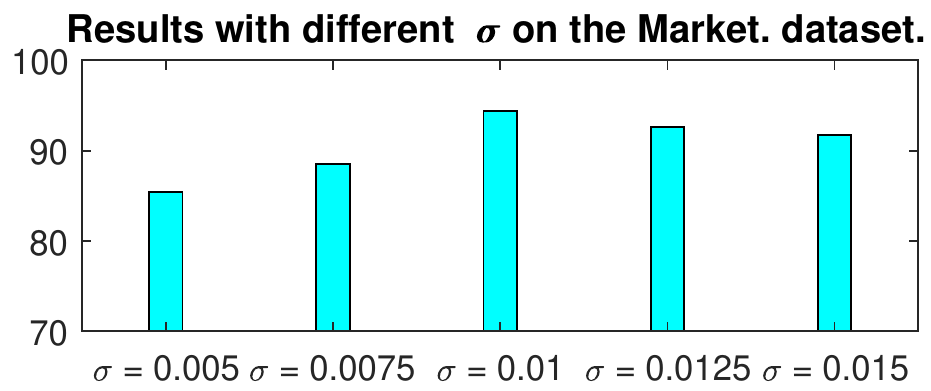}&
		\hspace{-0.5cm}
		\includegraphics[height = 2.1cm, width = 4.5cm]{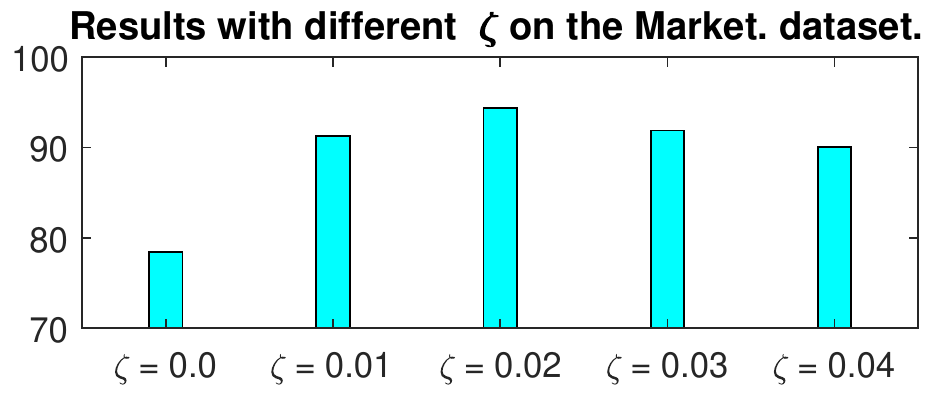}\\
		\hspace{-0.3cm}
		\includegraphics[height = 2.1cm, width = 4.5cm]{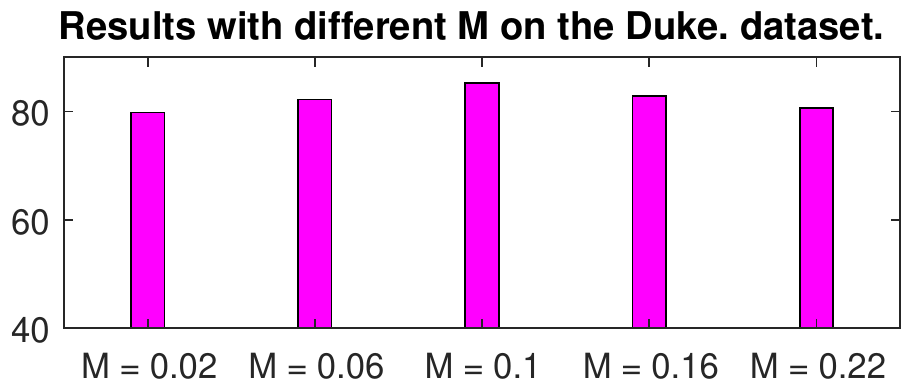}&
		\hspace{-0.5cm}
		\includegraphics[height = 2.1cm, width = 4.5cm]{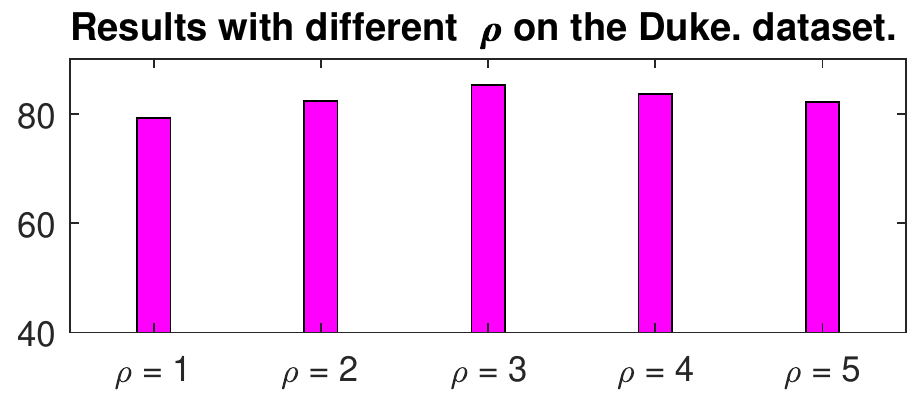}&
		\hspace{-0.5cm}
		\includegraphics[height = 2.1cm, width = 4.5cm]{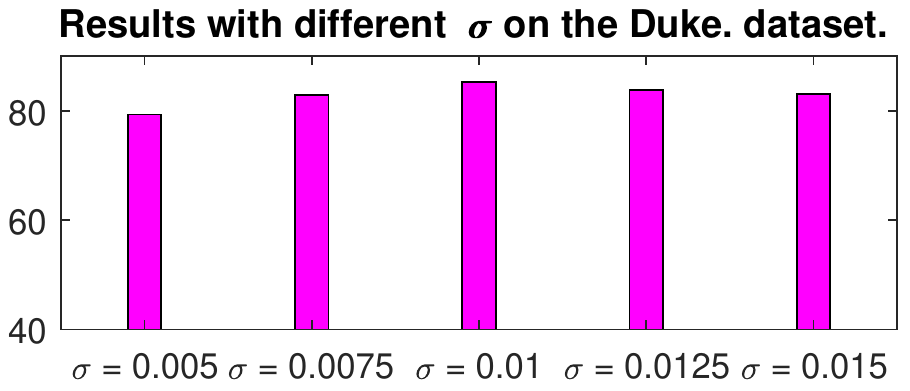}&
		\hspace{-0.5cm}
		\includegraphics[height = 2.1cm, width = 4.5cm]{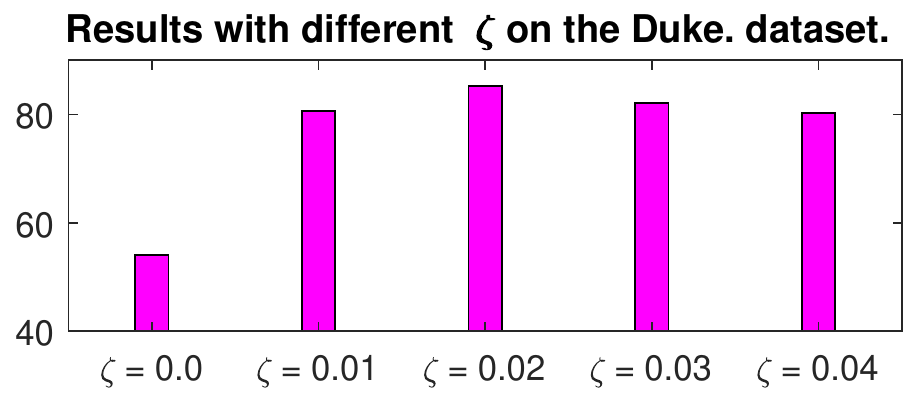}
	\end{tabular}
	\vspace{-0.1cm}
	\caption{Comparison of different parameter settings to the final person Re-ID performances on the six benchmark datasets. Specifically, the first to sixth rows show the detailed results on the 3DpeS, VIPeR, CUHK01, CUHK03, Market1501 and DukeMTMC-reID datasets, respectively.}
	\label{fig_6}
	\vspace{-0.3cm}
\end{figure*}

\begin{figure}[t]
	\centering
	\begin{tabular}{c}
		\includegraphics[height = 6.5cm, width = 8.5cm]{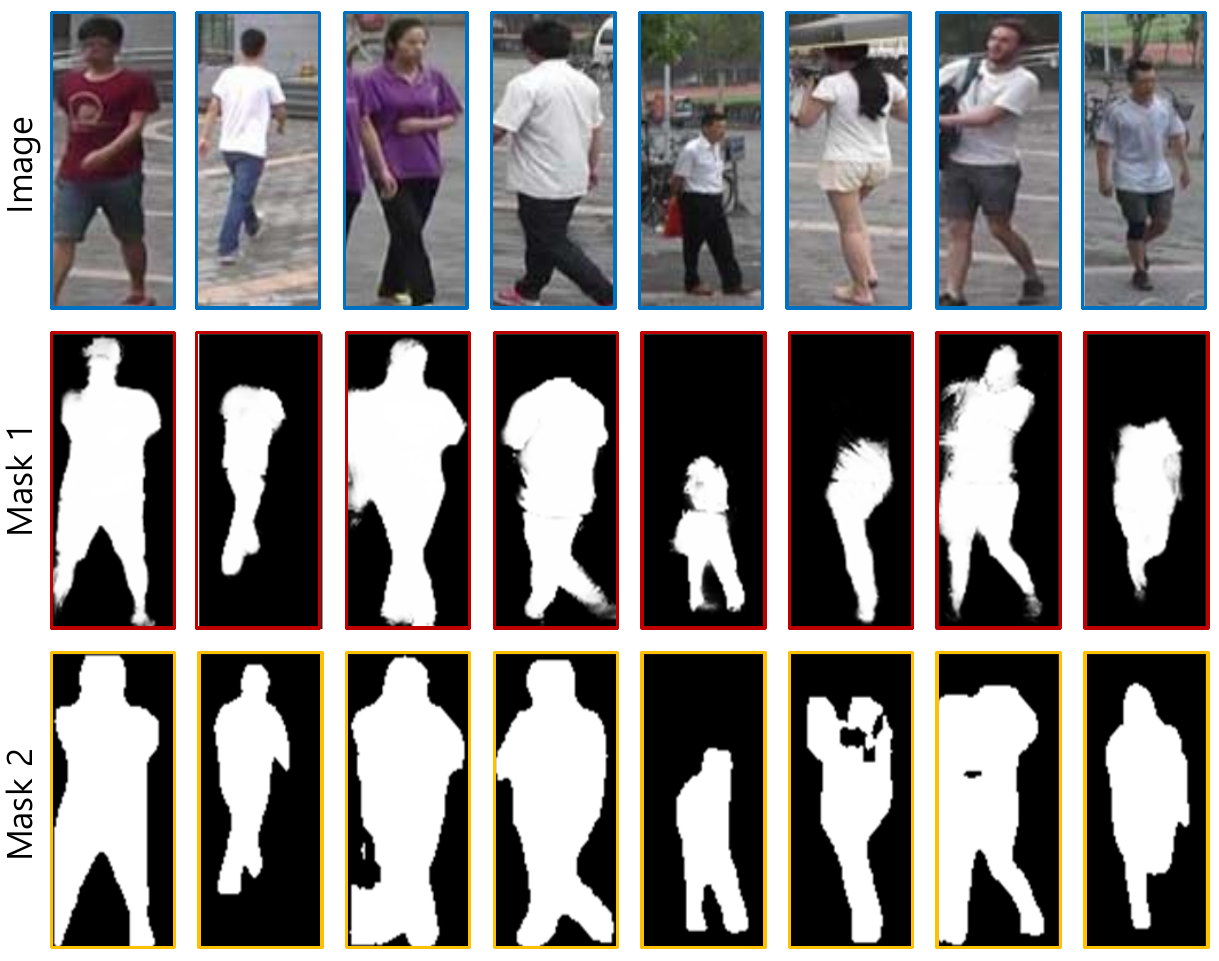}
	\end{tabular}
	\vspace{-0.1cm}
	\caption{Illustration of the generated masks by different methods, in which the first row shows the RGB images, the second row shows the masks generated by the saliency detection based method, and the third row shows the masks generated by the instance segmentation based method.}
	\label{fig_8}
	\vspace{-0.3cm}
\end{figure}

In addition, our FANN is robust to the quality of generated masks in a certain extent. In order to support this point of view, we take two methods, {\em i.e.}, one instance segmentation based method~\cite{Zheng_Jayasumana_Romera:2015} and one saliency detection based method~\cite{Deng_Hu_Zhu:2018}, to generate the ground truth masks. Some generated masks on the Market1501 dataset are shown in Fig.~\ref{fig_8}, in which we can see that masks generated by the instance segmentation based method are in higher quality than that by the saliency detection based method. Using the two kinds of masks as ground truth, we evaluate our method on the six benchmark datasets. The results are shown in Table~\ref{tab_8}, in which the `Baseline' denotes the result without using the mask information, `Mask 1' indicates the results by using the masks generated by the saliency detection based method, and `Mask 2' represents the results by using the masks generated by the instance segmentation based method. From the results, we can conclude that: 1) The two kinds of masks can help our FANN improves the person Re-ID results; 2) The poorer masks will lead to a bit worser results, however we also notice that the differences are not very large.

{\bf Robustness to parameter setting.} There are several hyper parameters in our method, and they are sensitive to the final person Re-ID results in different extents. In the following paragraphs, we will first evaluate our method with varying parameters: the margin parameter $\mathrm{M}$, the kernel parameters $\rho$ and $\sigma$, the weight parameter $\zeta$, and the initial adaptive weights $u$ and $v$. Specifically, we change one parameter and keep the others fixed in each experiment, so as to illustrate the sensitiveness of method to each parameter. Then, we further study how to set the number of residual blocks in the experiments on each dataset.

Firstly, we evaluate the influence of margin parameter $\mathrm{M}$, kernel parameters $\rho$ and $\sigma$, and weight parameter $\zeta$ to the final person Re-ID results. The detailed results are shown in Fig.~\ref{fig_6}, in which we have reported the Top 1 accuracy with different parameter settings. From the results, we can conclude three conclusions: 1) For the margin parameter, small $\mathrm{M}$ will lead to small discriminative power between the positive and negative pairs, and large $\mathrm{M}$ will make the model pay more attention to the hard training samples. Both parameter settings are not beneficial to keep a better generation ability of learned model on the testing data. What's more, there are a relative large sliding interval of $\mathrm{M}$ to keep an approximate performance on the testing data, when an optimal margin parameter is chosen in the experiments. 2) For the kernel parameters, small $\rho$ will make the reconstruction task sensitive to the isolated regions in the ground truth mask, and large $\rho$ will cause the edge blur when reconstructs the foreground mask. Small $\sigma$ will lead the reconstruction task sensitive to the difference between the reconstructed mask and ground truth mask, and large $\sigma$ will make the algorithm pay less attention to the foregrounds of input images. We argue that too small or too large $\rho$ and $\sigma$ are not beneficial to the generation ability of learned model, however our algorithm allows a large variation of the two parameters around the optimal values. For the weight parameter, small $\zeta$ will also make our algorithm pay less attention to the foregrounds of input images, and large $\zeta$ will make our algorithm pay too much attention to the foregrounds of input images. Therefore, an suitable weight should be chosen to keep the person Re-ID performance.

The main difference between our symmetric triplet loss function and the asymmetric triplet loss function is that it introduces another negative distance between two samples from the same camera view to regularize the gradient back-propagation, as shown in Fig.~\ref{fig_3}. As a result, the intra-class distance can be minimized and the inter-class distance can be maximized in each triplet unit, simultaneously. In our formulation, two negative distances are weighted to represent the inter-class distance, and the weight parameters can be adaptively updated in the feature learning process, so as to deduce the symmetric gradient back-propagation. Because we apply the L2 normalization after the resulting feature vectors, the final person Re-ID performance is very stable with different initializations to the weight parameters $u$ and $v$. In Table~\ref{tab_6}, we give some detailed analysis results on the six benchmark datasets. From the results, we can conclude that the Top 1 accuracy of our method on the six benchmark datasets are robust to different weight initializations. There are two underlying reasons: 1) The weight updating algorithm in Eq.~\eqref{eq_8} to  Eq.~\eqref{eq_10} can measure the difference between $\mathrm{d}(\mathbf{x}_i^1,\mathbf{x}_i^3)$ and $\mathrm{d}(\mathbf{x}_i^2,\mathbf{x}_i^3)$, so as to keep $\mathrm{d}(\mathbf{x}_i^1,\mathbf{x}_i^3)\approx \mathrm{d}(\mathbf{x}_i^2,\mathbf{x}_i^3)$ in the optimization. Therefore, the weights can be adaptively updated to keep the symmetric gradient back-propagation. 2) For the $i^{th}$ triplet input, the L2 normalization is applied to keep $\|\phi(\mathbf{x}_i^j, \mathbf{\Omega}_t)\|_2^2 = 1, j=1,2,3$ at the output layer, therefore the difference between $\mathrm{d}(\mathbf{x}_i^1,\mathbf{x}_i^3)$ and $\mathrm{d}(\mathbf{x}_i^2,\mathbf{x}_i^3)$ is bounded in $[0,2]$. As a result, the L2 normalization will make our algorithm more robust to the numerical stability. For comparison, we evaluate the performances of our method without using L2 normalization, as shown in Table~\ref{tab_7}, on the six benchmark datasets. From the results, we can see that: 1) The best results of two cases are similar on the six benchmark datasets, which indicates that both the unit sphere space and the Euclidean space are suitable for similarity comparison.\footnote{For fair comparison, we set the $\mathrm{M}=1.0$ when conduct experiments without using L2 normalization on the six benchmark datasets.} 2) Because the distances in unit sphere space are bounded, the performances on the six benchmark datasets are more stable even with different initializations.

\begin{table}[t]
	\caption{The matching rate (\%) on six benchmark datasets in term of $\mathbf{u}$ and $\mathbf{v}$ with using the L2 normalization.}
	\vspace{-0.1cm}
	\begin{center}
		\label{tab_6}
		\begin{tabular}{| c | c | c | c | c | c | c |}
			\hline
			\multicolumn{1}{|c|}{\multirow{2}{*}{Datasets}} &
			\multicolumn{2}{c|}{u = 1.0,v = 0.0} &
			\multicolumn{2}{c|}{u = 0.6,v = 0.4} &
			\multicolumn{2}{c|}{u = 0.4,v = 0.6}\\
			\cline{2-7}
			&Top 1 & Top 5 & Top 1  & Top 5 & Top 1 & Top 5\\
			\hline
			\hline
			3DPeS                            & 74.2 & 91.2 & \textbf{78.9} & \textbf{92.3} & 77.2 & 92.1\\
			VIPeR                            & 53.2 & 81.3 & \textbf{58.4} & \textbf{83.7} & 57.3 & 82.4\\
			CUHK01                           & 92.1 & 96.8 &   98.1 & \textbf{99.8} & \textbf{98.3} & 99.5\\
			CUHK03                           & 80.9 & 94.9 & \textbf{92.3} & 99.2 & 92.3 & \textbf{99.3}\\
			Market.                          & 88.9 & 92.1 & \textbf{94.4} & \textbf{95.3} & 92.1 & 94.8\\
			Duke.                            & 78.9 & 87.2 & \textbf{85.2} & \textbf{91.6} & 83.6 & 90.3\\
			\hline
		\end{tabular}
	\end{center}
	\vspace{-0.3cm}
\end{table}

\begin{table}[t]
	\caption{The matching rate (\%) on six benchmark datasets in term of $\mathbf{u}$ and $\mathbf{v}$ without using the L2 normalization.}
	\vspace{-0.1cm}
	\begin{center}
		\label{tab_7}
		\begin{tabular}{| c | c | c | c | c | c | c |}
			\hline
			\multicolumn{1}{|c|}{\multirow{2}{*}{Datasets}} &
			\multicolumn{2}{c|}{u = 1.0,v = 0.0} &
			\multicolumn{2}{c|}{u = 0.6,v = 0.4} &
			\multicolumn{2}{c|}{u = 0.4,v = 0.6}\\
			\cline{2-7}
			&Top 1 & Top 5 & Top 1  & Top 5 & Top 1 & Top 5\\
			\hline
			\hline
			3DPeS                            & 73.8 & 90.8 & \textbf{78.1} & \textbf{93.1} & 74.6 & 92.3\\
			VIPeR                            & 53.6 & 81.4 & \textbf{57.8} & \textbf{83.8} & 55.1 & 81.9\\
			CUHK01                           & 91.7 & 96.5 & \textbf{97.2} & \textbf{99.3} & 93.6 & 98.9\\
			CUHK03                           & 79.3 & 94.5 & \textbf{91.9} & \textbf{98.7} & 85.8 & 96.9\\
			Market.                          & 86.5 & 91.1 & \textbf{92.8} & \textbf{95.1} & 88.2 & 92.2\\
			Duke.                            & 77.1 & 86.5 & \textbf{84.1} & \textbf{90.6} & 78.9 & 88.5\\
			\hline
		\end{tabular}
	\end{center}
	\vspace{-0.3cm}
\end{table}

\begin{table}[t]
	\caption{The matching rate (\%) of using different numbers of residual blocks on the six benchmark datasets.}
	\vspace{-0.1cm}
	\begin{center}
		\label{tab_9}
		\begin{tabular}{| c | c | c | c | c | c |}
			\hline
			Datasets & num = 1 & num = 2 & num = 3 & num = 4 & num = 5 \\
			\hline
			\hline
			3DPeS                            & 77.1 & \textbf{78.9} & 76.2 & 74.1 & 69.1\\
			VIPeR                            & 56.2 & \textbf{58.4} & 54.2 & 49.1 & 43.6\\
			CUHK01                           & 94.9 & 97.2 & \textbf{98.1} & 96.5 & 91.0\\
			CUHK03                           & 82.3 & 87.1 & 89.4   & \textbf{92.3} & 91.4 \\
			Market.                          & 85.1 & 90.3 & 91.5   & \textbf{94.4} & 93.9 \\
			Duke.                            & 73.9 & 78.1 & 82.8 & \textbf{85.2} & 84.3\\
			\hline
		\end{tabular}
	\end{center}
	\vspace{-0.3cm}
\end{table}

In order to obtain better performances on both the small and large datasets, we use different numbers of residual blocks in the body part subnetwork. Specifically, we use two residual blocks on the 3DPeS and VIPeR datasets, three residual blocks on the CUHK01 dataset, and four residual blocks on the CUHK03, Market1501 and DukeMTMC-reID datasets, respectively. The main reason is that training a deep neural network usually needs a large amount of labeled data. Therefore, the under-fitting problem will be caused if we use a small dataset to train a complex network. We think this is the right reason why the deep neural networks, such as the VGGNet~\cite{Simonyan_Zisserman:2014} and ResNet~\cite{He_Zhang_Ren:2016}, usually can't work well on the small datasets. Besides, the heavy deep neural networks usually have stronger representation capacity than the light ones. As a result, we choose different numbers of residual blocks to adapt the scale of different datasets. In order to support our understanding, we evaluate the performance of our FANN with different numbers of residual blocks on each dataset. The results are shown in Table~\ref{tab_9}, in which we can find that the shallower networks can obtain better results than the deeper ones on the 3DPeS, VIPeR and CUHK01 datasets, while the deeper networks can obtain better results than the shallower ones on the CUHK03, Market1501 and DukeMTMC-reID datasets.

\begin{figure*}[t]
	\centering
	\begin{tabular}{c}
		\includegraphics[height = 6.2cm, width = 17.5cm]{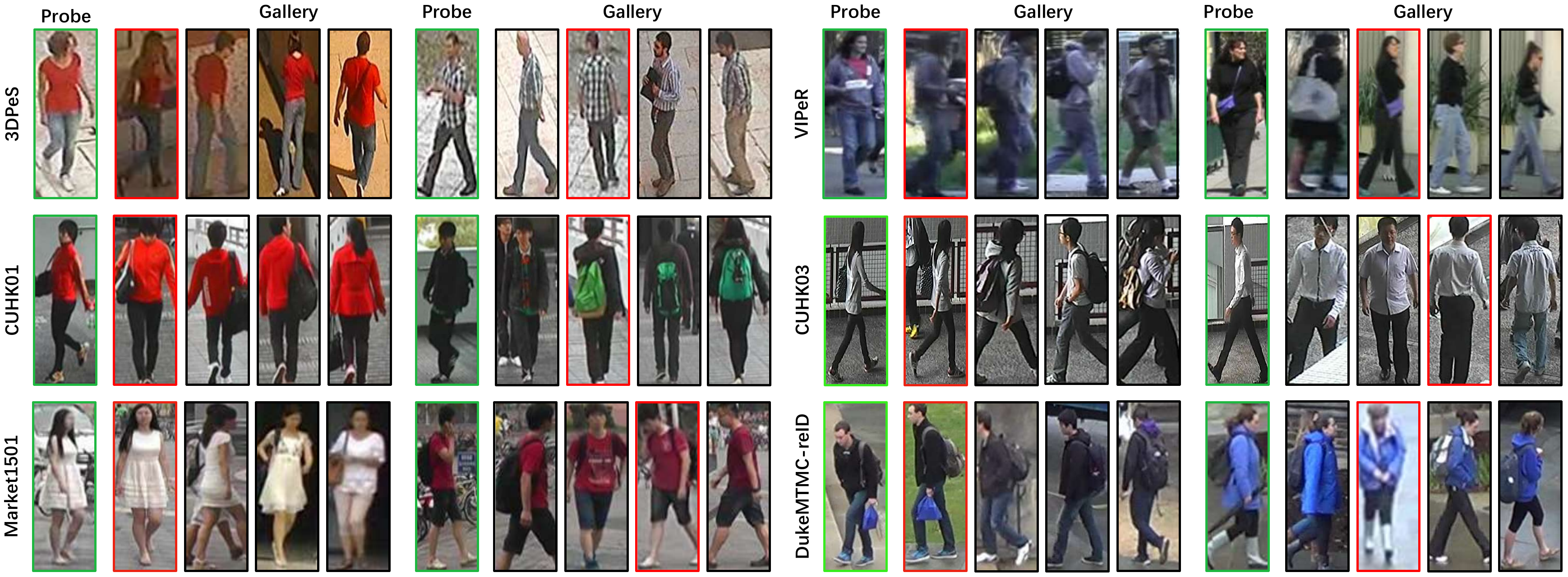}
	\end{tabular}
	\vspace{-0.1cm}
	\caption{Illustration of some ranking examples on the six benchmark datasets, in which the probe images are denoted by green bounding boxes and the matched references are indicated by using the red bounding boxes. Notice that both the successful cases and failure cases are given for better comparison.}
	\label{fig_7}
	\vspace{-0.3cm}
\end{figure*}

{\bf Some examples of ranking.} Finally, we illustrate some real ranking examples of our method on the six benchmark datasets, as shown in Fig.~\ref{fig_7}, including both the successful cases and failure cases. Specifically, the images in green boxes are the probes, which are used to find out the matched references from the gallery. The image in red box is a true match to the corresponding probe, in which the smaller order indicates the better performance. In the successful cases of our method, all the matched candidates are found out in the first place from various candidates in the gallery. The results indicate that our method can learn a discriminative feature representation to overcome the large cross-view appearance variations. However, we also notice that there are a fraction of failure cases in our method, in which the matched references can not be ranked firstly from the very similar candidates. In the future study, we will study how to enrich the diversity of similar training samples, so as to reduce the failure cases in solving person Re-ID problem.

\section{Conclusion}
\label{sec_con}
In this paper, we propose a simple yet effective deep neural network to learn a discriminative feature representation from the foreground of each input image for person Re-ID. Firstly, a FANN is constructed to jointly enhance the positive influences of foregrounds and weaken the side effects of backgrounds, in which an encoder and decoder network is built to guide the whole network to directly learn a discriminative feature representation from the foreground persons. Secondly, a novel local regression loss function is designed to deal with the isolated regions in the ground truth masks by considering the local information in a neighborhood. Thirdly, a symmetric triplet loss function is introduced to supervise the feature learning process, which can jointly minimize the intra-class distance and maximize the inter-class distance in each triplet unit. Extensive experiments on the 3DPeS, VIPeR, CUHK01, CUHK03, Market1501 and DukeMTMC-reID datasets are conducted, and the results have shown that our method can significantly outperform the state-of-the-art approaches.

\section*{Acknowledgment}
This work is jointly supported by the National Key Research and Development Program of China under Grant No. 2017YFA0700800, and the National Natural Science Foundation of China Grant No. 61629301.

\ifCLASSOPTIONcaptionsoff
  \newpage
\fi

{
\small
\bibliographystyle{IEEEtran}
\bibliography{FANN}
}

\begin{IEEEbiography}[{\includegraphics[width=1in,height=1.25in,clip]{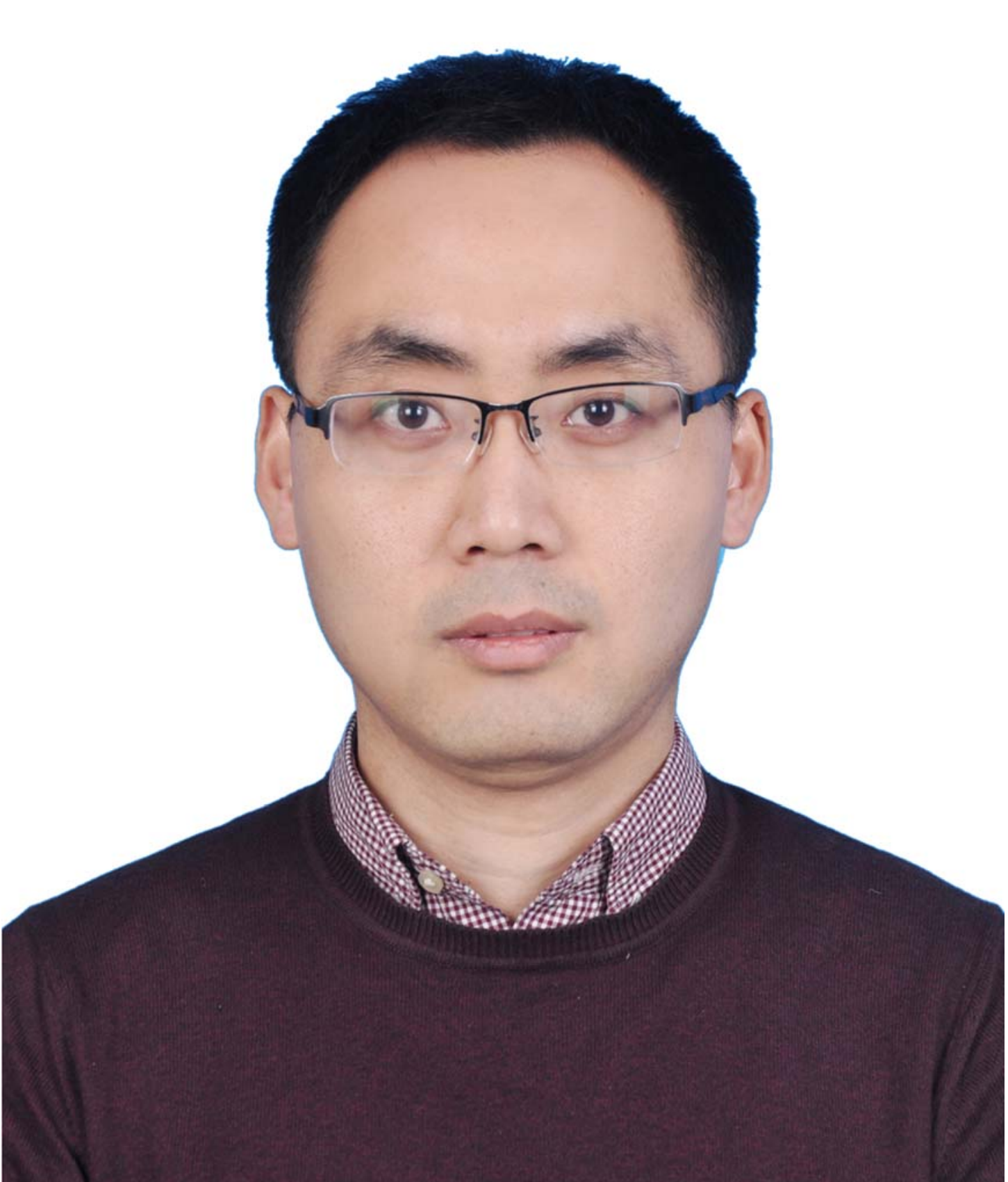}}]{Sanping Zhou}
received the M.E. degree from Northwestern Polytechnical University, Xi'an, China, in 2015. He is currently pursuing the Ph.D. degree in Institute of Artificial Intelligence and Robotics at Xi'an Jiaotong University. His research interests include machine learning, deep learning and computer
vision, with a focus on medical image segmentation, person re-identification, image retrieval, image classification and visual tracking.
\end{IEEEbiography}
\begin{IEEEbiography}[{\includegraphics[width=1in,height=1.25in,clip]{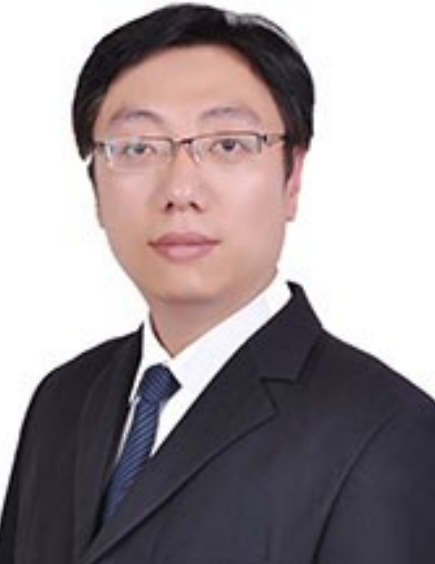}}]{Jinjun Wang}
received the B.E. and M.E. degrees from the Huazhong University of Science and Technology, China, in 2000 and 2003, respectively. He received the Ph.D. degree from Nanyang Technological University, Singapore, in 2006. From 2006 to 2009, he was with NEC Laboratories America, Inc., as a Research Scientist, and Epson Research and Development, Inc., as a Senior Research Scientist, from 2010 to 2013. He is currently a Professor with Xi'an Jiaotong University. His research interests include pattern classification, image/video enhancement and editing, content-based image/video annotation and retrieval, semantic event detection, etc.
\end{IEEEbiography}
\begin{IEEEbiography}[{\includegraphics[width=1in,height=1.25in,clip]{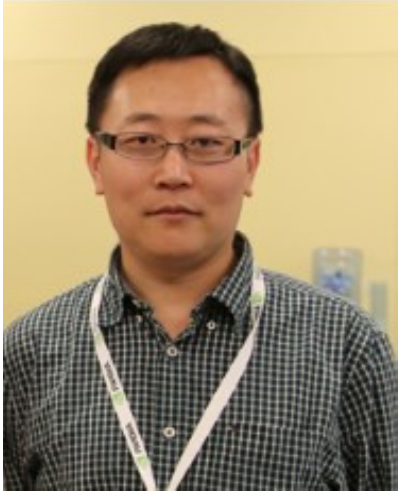}}]{Deyu Meng}
Deyu Meng received the B.S., M.S., and Ph.D. degrees from Xian Jiaotong University, Xian, China, in 2001, 2004, and 2008, respectively. From 2012 to 2014, he took his two-year sabbatical leave in Carnegie Mellon University. He is currently a Professor with the Institute for Information and System Sciences, School of Mathematics and Statistics, Xian Jiaotong University. His current research interests include self-paced learning, noise modeling, and tensor sparsity.
\end{IEEEbiography}
\begin{IEEEbiography}[{\includegraphics[width=1in,height=1.25in,clip]{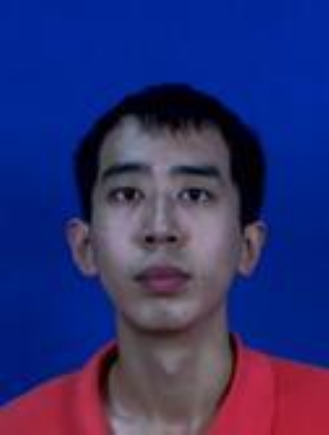}}]{Yudong Liang}
received the B.S. and Ph.D. degrees from Xi'an Jiaotong University, Xi'an, China, in 2010 and 2017, respectively. He is currently an assistant Professor with School of Computer and Information Technology, Shanxi University. His research interests include Machine Learning and Computer Vision, with a focus on image super-resolution, image quality assessment and deep learning.
\end{IEEEbiography}
\begin{IEEEbiography}[{\includegraphics[width=1in,height=1.25in,clip]{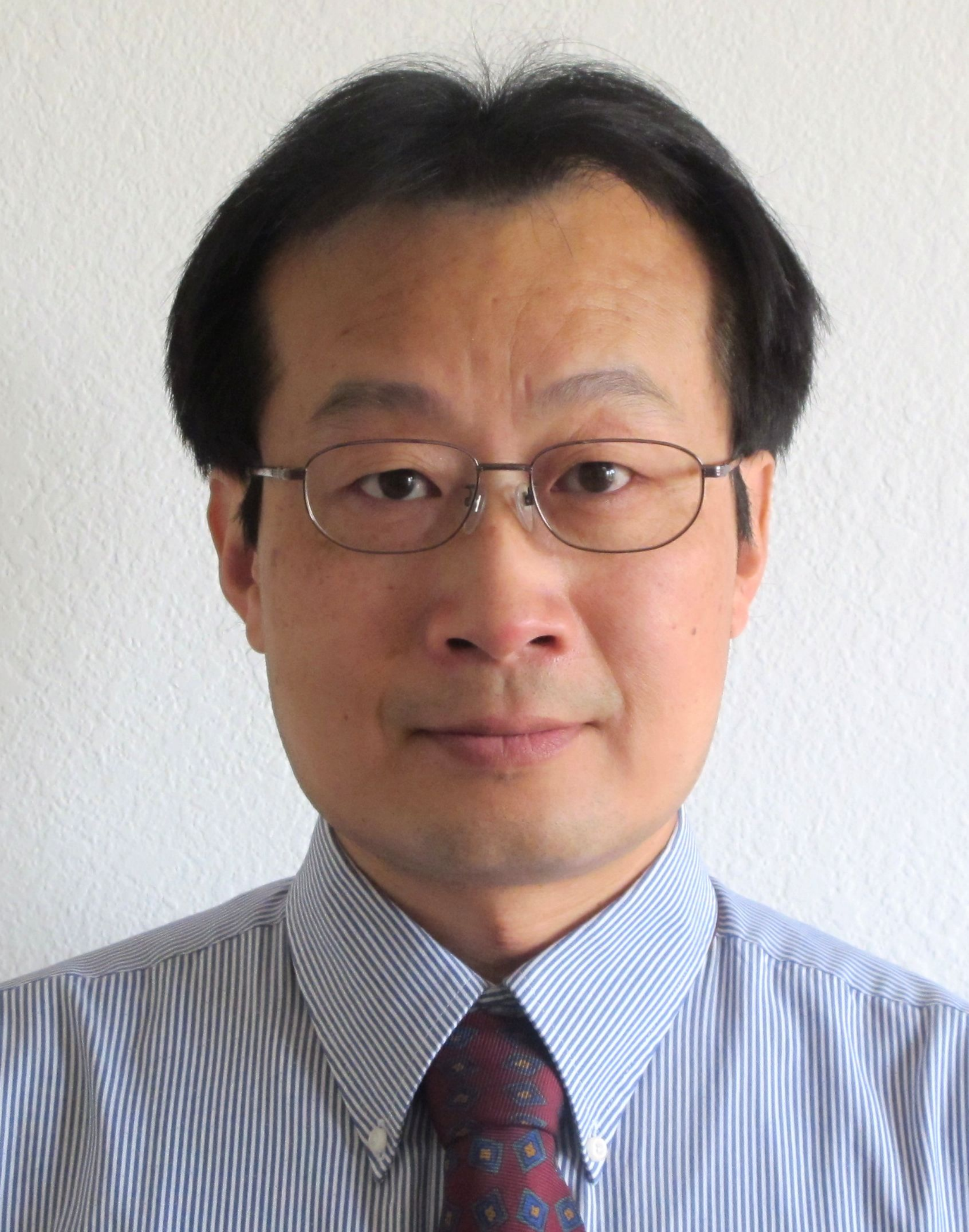}}]{Yihong Gong}
received the B.S., M.S., and Ph.D. degrees in Electrical Engineering from The University of Tokyo, Japan, in 1987, 1989, and 1992, respectively. In 1992, he joined Nanyang Technological University, Singapore, as an Assistant Professor with the School of Electrical and Electronic Engineering. From 1996 to 1998, he was a Project Scientist with the Robotics Institute, Carnegie Mellon University, USA. Since 1999, he has been with the Silicon Valley branch, NEC Labs America, as a Group Leader, the Department Head, and the Branch Manager. In 2012, he joined Xi'an Jiaotong University, China, as a Distinguished Professor. His research interests include image and  video  analysis,  multimedia  database  systems, and machine learning.
\end{IEEEbiography}
\begin{IEEEbiography}[{\includegraphics[width=1in,height=1.25in,clip]{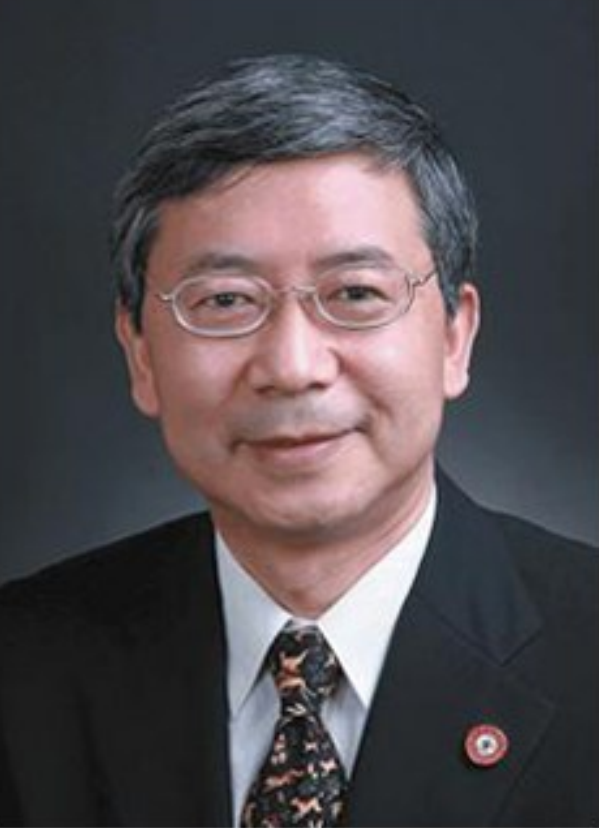}}]{Nanning Zheng}
(SM93-F06) graduated from the Department of Electrical Engineering, Xian Jiaotong University, Xian, China, in 1975, and received the M.S.  degree in information and control engineering from Xian Jiaotong University in 1981 and the Ph.D. degree in electrical engineering from Keio University, Yokohama, Japan, in 1985. He jointed Xian Jiaotong University in 1975, and he is currently a Professor and the Director of the Institute of Artificial Intelligence and Robotics, Xian Jiaotong University. His research interests include computer vision, pattern recognition and image processing, and hardware implementation of intelligent systems. Dr. Zheng became a member of the Chinese Academy of Engineering in 1999, and he is the Chinese Representative on the Governing Board of the International Association for Pattern Recognition.
\end{IEEEbiography}

\end{document}